\documentclass[journal]{IEEEtran}
\usepackage{amsmath,amsfonts}
\usepackage{algorithmic}
\usepackage{algorithm}
\usepackage{array}
\usepackage[caption=false,font=normalsize,labelfont=sf,textfont=sf]{subfig}
\usepackage{textcomp}
\usepackage{stfloats}
\usepackage{url}
\usepackage{verbatim}
\usepackage{graphicx}
\usepackage{cite}
\usepackage{amsthm}
\usepackage{amssymb}
\usepackage{xcolor}

\usepackage{multirow} 
\usepackage{tabularx}
\usepackage{array}
\usepackage{makecell}

\usepackage{newtxtext}
\usepackage{hyperref}

\begin{document}

\title{Reinforcement Learning with Knowledge Representation and Reasoning: A Brief Survey}

\author{Chao Yu, Shicheng Ye and Hankz Hankui Zhuo

        
\thanks{All authors are affiliated with the School of Computer Science and Engineering at Sun Yat-sen University. The corresponding author's email is yuchao3@mail.sysu.edu.cn.
}
\thanks{We gratefully acknowledge support from the National Natural
Science Foundation of China (No. 62076259), the Fundamental
and Applicational Research Funds of Guangdong Province (No.
2023A1515012946), and the Fundamental Research Funds for the
Central Universities-Sun Yat-sen University. This research was supported by Meituan.}}

\markboth{Journal of \LaTeX\ Class Files,~Vol.~14, No.~8, August~2021}%
{Shell \MakeLowercase{\textit{et al.}}: A Sample Article Using IEEEtran.cls for IEEE Journals}


\maketitle

\begin{abstract}
\emph{Reinforcement Learning} (RL) has achieved tremendous development in recent years, but still faces significant obstacles in addressing complex real-life problems due to the issues of poor system generalization, low sample efficiency as well as safety and interpretability concerns. The core reason underlying such dilemmas can be attributed to the fact that most of the work has focused on the computational aspect of value functions or policies using a representational model to describe atomic components of rewards, states and actions etc, thus neglecting the rich high-level declarative domain knowledge of facts, relations and rules that can be either provided a priori or acquired through reasoning over time. Recently, there has been a rapidly growing interest in the use of \emph{Knowledge Representation and Reasoning} (KRR) methods, usually using logical languages, to enable more abstract representation and efficient learning in RL. In this survey, we provide a preliminary overview on these endeavors that leverage the strengths of KRR to help solving various problems in RL, and discuss the challenging open problems and possible directions for future work in this area.
\end{abstract}


\section{Introduction}
\IEEEPARstart{R}{einforcement} learning (RL)~\cite{sutton2018reinforcement} has made tremendous theoretical and technical breakthroughs in the past few years, achieving great successes in a wide range of domains, such as game AI design, robotics control, healthcare treatment optimization, and autonomous driving, just to name a few~\cite{li2017deep}. Despite such progresses, it is believed that there is still a \textit{theory-to-application gap} in the RL research that hinders the wide deployment of RL techniques in real-world problems. The major reasons can be attributed to the following aspects:  \emph{the low sample efficiency} during the exploration process, i.e., millions of interactions are usually required even for relatively simple problems~\cite{yang2021exploration}, \emph{the poor policy generalization capabilities} due to the sim-to-real or cross domain difference during transfer learning~\cite{kirk2021survey}, and  \emph{the lack of consideration of critical concerns such as safety, interpretability and robustness}~\cite{garcia2015comprehensive}. These problems become even more notable in complex environments characterized by sparse rewards, partial observations and high dynamics caused by other co-learners in a multiagent system.

Recently, there have been rapidly growing interests in the use of methods and tools from the area of \emph{Knowledge Representation and Reasoning} (KRR)~\cite{brachman2004knowledge} to help solving RL problems. The basic idea is to abstract the representation in RL by formal languages such that the problems can be described more compactly and thus solved more efficiently and transparently. 
In this way, rich declarative domain knowledge of objects and their relations, either provided a priori or acquired through reasoning over time, can be readily incorporated into the learning process in order to further improve the learning performance of RL. 
Until now, an increasing number of methods have been proposed in the literature, using a wide variety of representational languages and reasoning techniques in the realm of KRR, to scale up RL to large domains that are prohibitive for the traditional RL methods. 

In specific, the use of KRR methods in RL offers some prominent advantages. \textbf{First}, traditional representations in RL generally require all possible states/actions to be represented explicitly, usually using atomic attribute-values, thus are not able to represent the high-level structural relations of the problem. Formal languages-specified representations, in contrast, allow for a more general and intuitive way of specifying and using knowledge about a problem regarding \textit{objects with their attributes and relations}, \textit{rules with the manipulated domain dynamics}, and \textit{preconditions with their effects}, etc., thus inducing more powerful abstractions over the problem and potentially more efficient learning performance. \textbf{Second}, RL depends critically on the choice of reward functions to capture the desired behavior and constraints. Usually, these rewards are handcrafted by experts and can only represent heuristics for relatively simple tasks. 
Using simple policies or rewards expressed by formal languages, however, can easily encode human instructions or domain constraints into the policy, thus enabling of learning more complex behaviors, especially for tasks with multiple objectives, diverse constraints and complex relations.
\IEEEpubidadjcol
\textbf{Moreover}, the existing RL methods usually lack the ability to reason over high-level ideas or knowledge of \textit{facts} and \textit{beliefs} as most of these methods directly work on primitive components of rewards, actions and states etc. Therefore, it is difficult to implement rich cognitive functions such as \textit{transfer learning}, \textit{analogical analysis}, or \textit{rule-based reasoning}, which are significant abilities of human intelligence. Using  formal descriptive languages to represent the tasks can better facilitate transfer learning among different tasks (i.e., \textit{generalization}) when the series of tasks do not necessarily share the same structure, but instead are specified with high-level events. For example, consider a scenario when an agent has learned the task of ``\textit{delivering coffee and mail to office}''. When facing a new task of ``\textit{delivering coffee or mail to office}'', it is unclear how existing RL methods would model these as the same distribution of tasks and enable knowledge transfer among them. By exploiting task modularity and decomposition with higher abstraction, better generalization can be achieved by combining and reasoning over already known primitives (i.e., ``\textit{delivering coffee to office}'' and ``\textit{delivering mail to office}''). \textbf{Last but not the least}, the black-box training of the existing RL methods is largely opaque to the users, rendering the learned policies difficult to understand and trust, and thus unsuitable for safety-critical domains such as finance, healthcare and self-driving, where robustness, safety, verifiability and interpretability are the major concerns. In contrast, simple policies consisting of explicit symbolic expressions can facilitate human understanding, while increasing the transparency and explainability of the learning process and policies.

In this survey, we provide a preliminary overview on the rich literature that leverages the strengths of KRR to help addressing various problems in RL, based on the category of learning goals, i.e., \emph{the efficiency}, \emph{the generalization}, as well as \emph{the safety and interpretability}. In terms of \emph{efficiency}, the goal of using KRR methods is to encode prior knowledge into RL in order to improve the learning efficiency of traditional RL methods, either through specifying the tasks using increasingly expressive formal languages, or through combining high-level symbolic action models, as discussed in Section \ref{sec:three}. The \emph{generalization} category aims to employ KRR methods to generalize RL algorithms well to unseen situations at deployment time by avoiding overfitting to the training environment (i.e., \textit{the one-shot transfer learning problem}) or transferring knowledge learned from previous tasks to solve a series of new but related tasks quickly (i.e., \textit{the lifelong or continual learning problem}), which is addressed in Section \ref{sec:four}. 
The \emph{safety and interpretability} category, detailed in Section \ref{sec:five}, focuses on satisfying constraints, verifying models and providing explicit interpretations of the RL methods by using formal descriptive expressions and symbolic languages.  Finally, we also discuss the challenging open problems, and possible directions for future work in this interdisciplinary area in Section \ref{sec:sixth}.

\section{Background}

\begin{figure}[t]
    \centering
    \includegraphics[width=0.5\textwidth]{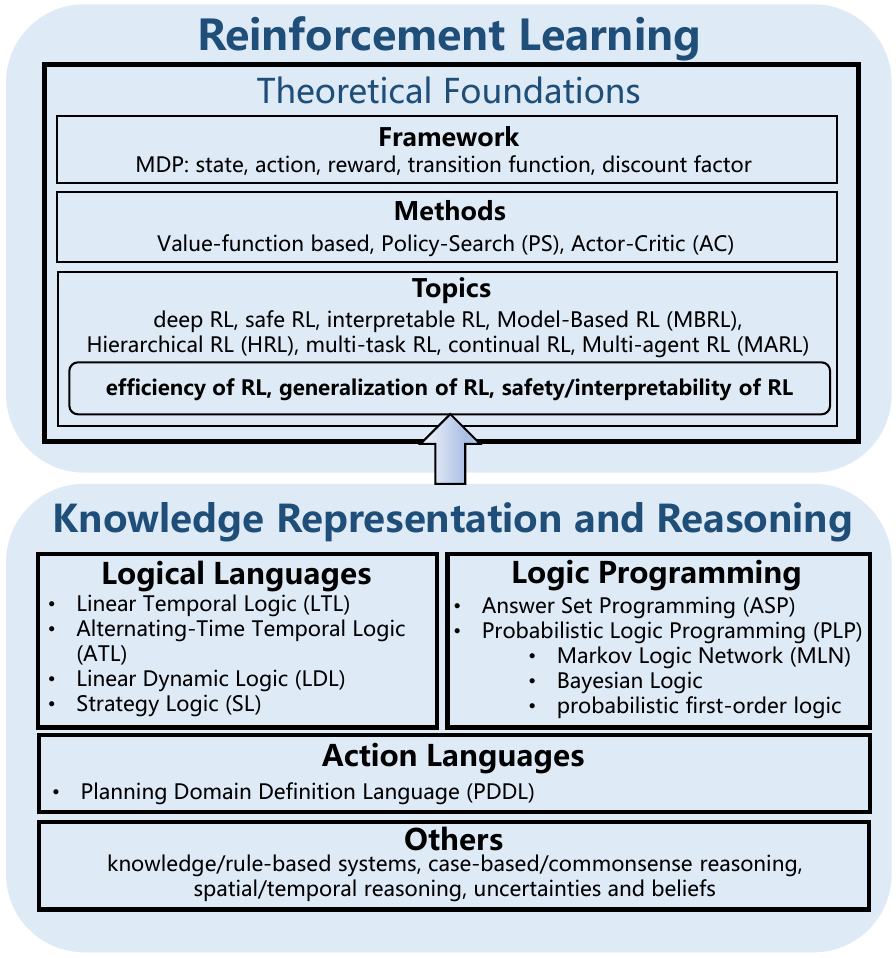}
    \caption{Key components and concepts in RL and KRR.}
    \label{fig:background}
\end{figure}

This section briefly introduces basic knowledge and key concepts related to KKR and RL that are the two main topics we consider in this survey. Fig.~\ref{fig:background} provides a summary of these two topics. 

\subsection{RL}

As a subfield of machine learning, RL provides a learning paradigm that enables an agent to learn effective strategies by trial-and-error interactions with the environment~\cite{sutton2018reinforcement}. The \emph{Markov Decision Process} (MDP) has been used as a general framework to formalize an RL problem by a 5-tuple $\mathcal{M} = (\mathcal{S}, \mathcal{A}, \mathcal{P}, \mathcal{R},\gamma)$, where $\mathcal{S}$ is a finite \emph{state} space, and $s_t\in \mathcal{S}$ denotes the state of an agent at time $t$; $\mathcal{A}$ is a set of \emph{actions} available to the agent, and $a_t\in \mathcal{A}$ denotes the action that the agent performs at time $t$; $\mathcal{P}(s,a,s^{\prime}): \mathcal{S} \times \mathcal{A} \times \mathcal{S} \rightarrow [0,1]$ is a Markovian \emph{transition function} when the agent transits from state $s$ to state $s^{\prime}$ after taking action $a$; $\mathcal{R}: \mathcal{S} \times \mathcal{A} \rightarrow \Re$ is a \emph{reward function} that returns the immediate reward $\mathcal{R}(s,a)$ to the agent after taking action $a$ in state $s$; and $\gamma \in [0,1]$ is a \emph{discount factor}. An agent's \emph{policy} $\pi: \mathcal{S} \times \mathcal{A} \rightarrow [0,1]$ is a probability distribution that maps an action $a \in \mathcal{A}$  to a state $s \in \mathcal{S}$. 
The expected reward of following a policy when starting in state $s$, $V^{\pi}(s)$, can be defined as follows:
\begin{equation}\label{equ:v}
 V^{\pi}(s) \triangleq E_{\pi} \left[\sum_{t=0}^{\infty}\gamma^{t}\mathcal{R}(s_{t},\pi(s_{t}))|s_{0}=s\right].
\end{equation}

The goal of an MDP problem is to compute an \emph{optimal policy} $\pi^{\ast}$ such that $V^{\pi^{\ast}}(s)\geq V^{\pi}(s)$ for every policy $\pi$ and every state $s\in S$.  
One of the most important and widely used RL approaches is Q-learning~\cite{watkins1992q}, with its one-step updating rule given by
\begin{equation}\label{equ:Q1}
 Q_{t+1}(s,a) = (1-\alpha_t)Q_{t}(s,a) + \alpha_{t}[R(s,a) + \gamma \max_{a^{\prime}} Q_t(s^{\prime},a^{\prime})],
\end{equation}
where  $\alpha_{t}\in(0,1]$ is a learning rate that controls the contribution of the new experience to the current estimate. Besides the \emph{value-function} based methods that maintain a value function whereby a policy can be derived, direct \emph{Policy-Search} (PS) algorithms  estimate the policy directly without representing a value function explicitly, whereas the \emph{Actor-Critic} (AC) methods keep separate, explicit representations of value functions and policies. The major theoretical research issues and subfield topics in RL include but are not limited to \textit{Deep RL} (DRL)~\cite{arulkumaran2017deep}, \textit{safe RL}~\cite{garcia2015comprehensive}, \textit{interpretable RL}~\cite{glanois2024survey}, \textit{Model-Based RL} (MBRL)~\cite{moerland2023model}, \textit{Hierarchical RL} (HRL)~\cite{pateria2021hierarchical}, \textit{multi-task RL}~\cite{vithayathil2020survey}, \textit{continual RL}~\cite{khetarpal2022towards}, \textit{Multi-Agent RL} (MARL)~\cite{zhang2021multi}, \textit{relational RL}~\cite{zambaldi2018relational}, and so on. A more comprehensive and in-depth review on these issues can be found in \cite{li2017deep,sutton2018reinforcement}.

\subsection{KRR}
KRR is a subarea of AI focusing on how information about the world can be represented formally and manipulated in an automated way by some reasoning programs~\cite{zhang2020survey,brachman2004knowledge}. 
 Research on KRR has a long history of more than half century along with the development of AI, covering various topics including \textit{knowledge representation languages}, \textit{knowledge/rule-based systems}, \textit{case-based/commonsense reasoning}, \textit{spatial/temporal reasoning}, \textit{action models}, and \textit{uncertainties and beliefs}, etc.	
Symbolic logic plays a vital role in KRR as logic provides a strict and formal representation of knowledge and their entailment relations. There are numerous logical languages to explicitly represent and reason over knowledge, mostly built upon the \textit{First Order Logic} (FOL), such as the \textit{Linear Temporal Logic} (LTL) \cite{pnueli1977temporal} and its various extensions to model time dependence of events, and the \textit{Strategy Logic} (SL)~\cite{mogavero2014reasoning} to capture strategic abilities of agents.
\textit{Logic Programming} (LP)~\cite{lloyd2012foundations} provides a universally quantified expression of facts and rules using a system of formal logic, while \textit{Probabilistic Logic Programming} (PLP)~\cite{de2015probabilistic}, such as \textit{Markov Logic Network} (MLN)~\cite{2006Markov},  combines probability with logic for dealing with more complex problems with uncertainty.
Last, \textit{Action Languages}, such as the \textit{Planning Domain Definition Language} (PDDL)~\cite{1998PDDL}, are used for  specifying state transition diagrams and formal models of the effects of actions on the world such that planning methods can then be applied to generate action sequences to achieve certain goals. Refer to~\cite{brachman2004knowledge}  for a more comprehensive review on these issues.

\section{KRR for \textit{Efficiency} of RL}
\label{sec:three}

Logically-specified representations allow for a more general and intuitive way of specifying and using knowledge about \textit{objects}, \textit{facts} and \textit{rules}, inducing powerful abstractions over the original problem. Encoding such prior knowledge is thus able to improve the learning efficiency of traditional RL methods, either through specifying the tasks using increasingly expressive formal languages and \textit{finite state machines} (FSAs), or combining high-level symbolic planning models with RL. 
KRR techniques aimed at improving the efficiency of RL are summarized in Table \ref{tab:efficiency}.

\subsection{Task Representation}

Typically, RL algorithms require the user to manually design a reward function that encodes the desired task, which might be tricky for complex, long-horizon tasks~\cite{jothimurugan2019composable}. Particularly, for tasks with multiple objectives and constraints, manually devising a single reward function that balances different parts of the task could be challenging. Moreover, different reward functions can encode the same task, and the choice of reward function can have a large impact on the final performance of the RL algorithm. 
The use of formal methods to synthesize reward functions is a promising approach to address the above issues.

\begin{figure}[t]
    \centering
    \includegraphics[width=0.5\textwidth]{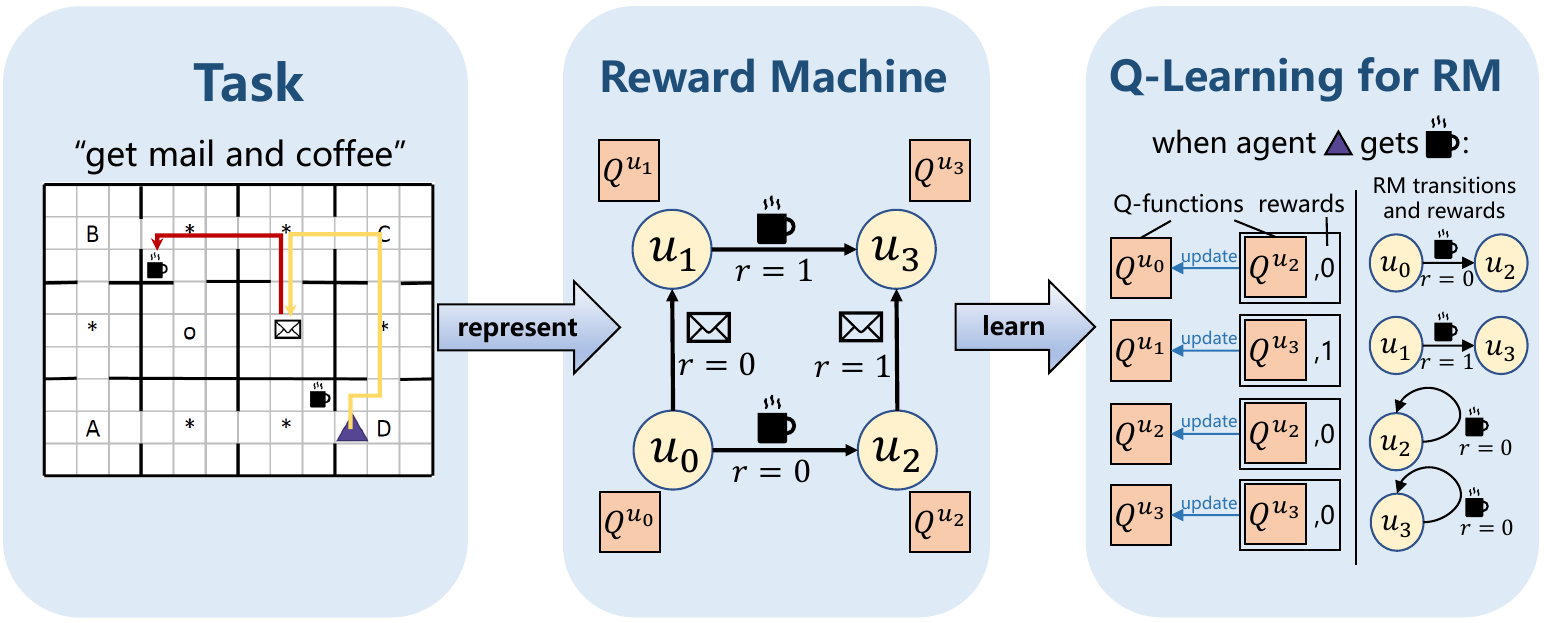}
    \caption{An illustration of representing a task as an RM and one-step update of the QRM algorithm. 
   The task of \textit{``get mail and coffee''} can be modeled using an RM with 4 abstract states $u_0,u_1,u_2$ and $u_3$. Starting at $u_0$ (implying the initial stage of the task), when an agent gets a coffee, the RM receives this abstracted description and then transitions to $u_2$ (implying the stage that the coffee has been gotten) and returns a reward $r=0$. When the RM reaches its accepting state $u_3$, it indicates that the task has been completed. 
    }
    \label{fig:QRM}
\end{figure}

\textbf{Reward Machines.}
Icarte et al.~\cite{icarte2018using} first propose \textit{Reward Machines} (RMs)-a type of FSA-that supports the specification of reward functions while supporting high-level task decomposition. 
An RM takes abstracted descriptions of the environment as input, and outputs rewards that guide the agent towards task completion. 
An abstract state in RMs is a temporally extended state of the environment, informing the agent which stage the task is currently in. 
When an RM receives the abstracted descriptions from the environment, it transitions to another state, which implies that the task goes to another stage. 
Each state $u_i$ is assigned with a Q-function $Q^{u_i}$ to learn its subpolicy.
A new algorithm called \emph{Q-Learning for Reward Machines} (QRM)~\cite{icarte2018using} is then proposed to appropriately decompose the RM and simultaneously learn subpolicies for different subtasks.
The rationale behind its off-policy learning mechanism lies in that the experiences gained by the agent while learning a specific subtask can be beneficial for developing the subpolicies of other subtasks.
As illustrated in Fig.~\ref{fig:QRM}, when an agent gets a coffee, for each abstract state $u$, all the Q-functions are updated according to the internal transitions and rewards of the RM simultaneously by
\begin{equation}
    Q^{u}(s,a)\leftarrow (1-\alpha)Q^{u}(s,a)+\alpha [r+\gamma \max_{a'} Q^{u'}(s',a')]
\end{equation}
where $r=\delta_r(u,l)$ and $u'=\delta_u(u,l)$ are the reward and the next state of $u$, and $l$ is the current abstracted descriptions of the environment. $\delta_r$ and $\delta_u$ are the reward function and the transition function of RM, respectively.
QRM is guaranteed to converge to an optimal policy, in contrast to HRL methods which might converge to suboptimal policies. 

In a following paper, Icarte et al.~\cite{icarte2022reward} propose variants of QRM including \textit{Hierarchical RM} (HRM) to learn policies for tasks specified using RMs. 
HRM maintains a set of \textit{options}, each of which corresponds to a transition in RM. 
HRM also learns a high-level policy to choose options  using $t$-step Q-learning, where $t$ is the number of steps that the current option lasts for. 
With the strength of learning policies for all options simultaneously, HRM is effective at learning good policies for a task specified by an RM. However, it might converge to suboptimal solutions even in simple tabular cases due to the myopic option-based approach.

\begin{figure}[t]
    \centering
    \includegraphics[width=0.48\textwidth]{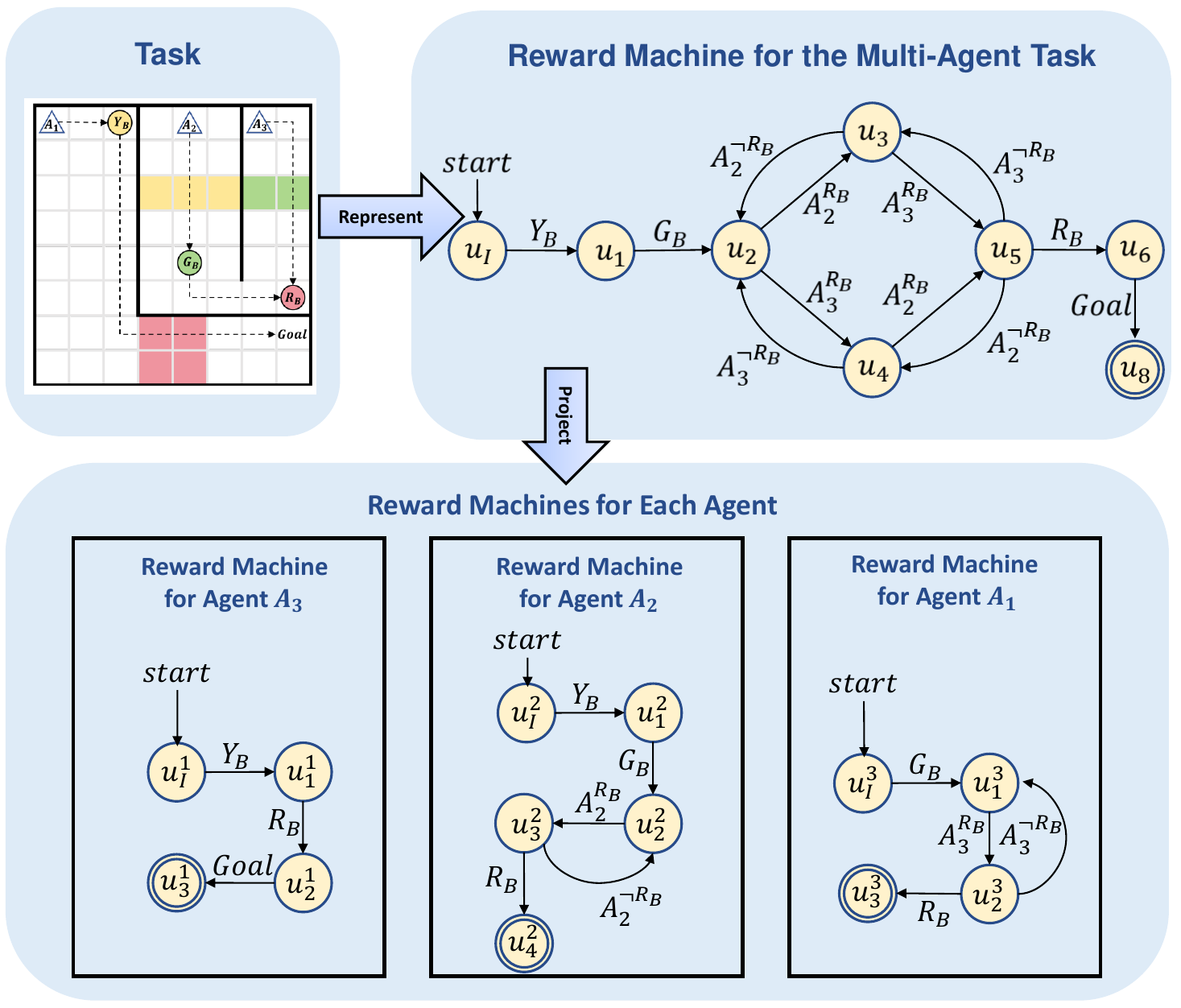}
    \caption{An illustration of handling a multi-agent cooperative task using RMs (adapted from~\cite{neary2020reward}). In the scenario, three agents must work collaboratively to guide agent $A_1$ to the target location $Goal$. The colored zones indicate regions where a corresponding color-coded button must be activated for an agent to pass. For the red zone, both agents $A_2$ and $A_3$ must press the red button simultaneously to allow agent $A_1$ to pass the red zone, while the yellow and green zones require only one agent to press the respective button. This multi-agent cooperative task can be represented using the corresponding RM, and the set of events of the RM is $\sum = \{Y_B, G_B, R_B, A_2^{R_B}, A_2^{\neg R_B}, A_3^{R_B}, A_3^{\neg R_B}, Goal\}$. For the RM representing the entire task, it can be decomposed into RMs corresponding to each agent's subtask by projecting onto the local event set of each agent.
    }
    \label{fig:rm4ma}
\end{figure}

There are some following studies that extend the RMs to encode a team task in cooperative MARL.
In specific, Neary et al.~\cite{neary2020reward} propose using RMs to explicitly encode the necessary interdependencies between teammates, allowing the team-level task specified by a RM to be decomposed into sub-tasks for individual agents, as shown in Fig.~\ref{fig:rm4ma}.
Hu et al.~\cite{hu2021decentralized} propose a framework, in which each agent not only observes its local state and RM state like~\cite{neary2020reward}, but also has access to the state information of its neighbors.
The algorithm conditions the truncated Q-function within a \textit{k}-hop neighborhood to tackle the increased complexity.
However, both works~\cite{neary2020reward,hu2021decentralized} have to rely on the assumption of weak interdependencies among the agents.
Instead, Zheng et al.~\cite{zheng2024multi} introduce an algorithm to handle complex scenarios with highly interdependent agents.
A complex task is decomposed into simpler subtasks using a hierarchical structure of RMs, where a higher-level RM policy selects lower-level RMs, and the lowest-level RM policies dictate the actions.
However, all these works assume that the multi-agent RM is pre-defined, requiring extensive human prior knowledge. 

To avoid the difficulties of manually constructing RMs, some studies focus on inferring RMs from the agent’s learning experience instead of having them specified by the user.
Icarte et al.~\cite{icarte2019learning} formulate the task of learning RMs as a discrete optimization problem, and propose an efficient local search approach to solve it.
Based on this work, they provide further details about this learning pipeline and propose three novel formulations to learn RMs~\cite{icarte2023learning}.
Xu et al.~\cite{xu2020joint} introduce an iterative algorithm for joint inference of RMs and policies in RL.
To obtain a correct RM, an agent refines a hypothesis RM using counterexamples, where the rewards deviate from those predicted by the current hypothesis RM.
Furelos-Blanco et al.~\cite{furelos2020induction} propose a method to use \textit{inductive logic programming} (ILP) to learn the subgoal automaton.
Although it does not address the strict RM construction problem, the constructed automaton resembles the RM in structure.
Furthermore, several subsequent studies have explored inferring RMs in multi-agent settings~\cite{varricchione2023synthesising, ardon2023learning}.
However, the above works tend to assume the availability of noise-free high-level proposition detectors, which can be challenging to obtain in practice.
There are some works that consider noisy interpretations of propositions~\cite{hatanaka2023reinforcement,li2024reward}, but their applicability is highly dependent on the performance of uncertain estimators.

\begin{figure}[t]
    \centering
    \includegraphics[width=0.48\textwidth]{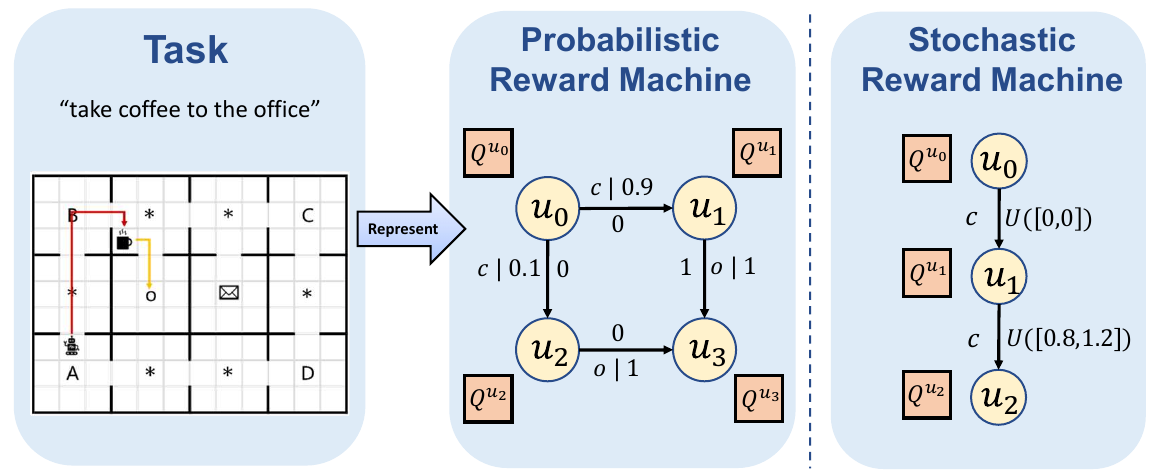}
    \caption{An illustration of the PRM and SRM. The original task is to bring coffee to the office, with a reward of 1 upon successful completion. To expand this into a scenario with random rewards, an element of randomness is introduced: the coffee machine has a 10\% chance of malfunctioning, resulting in substandard coffee, which means no reward can be obtained. 
    This task specification can be utilized to construct the corresponding PRM and SRM. In the PRM, a nondeterministic transition function is employed to achieve this randomness, where $y \overset{l|p}{\underset{r}{\rightarrow}} y\prime$ indicates that $y$ transitions to $y\prime$ on high-level label $l$ with probability $p$, receiving reward $r$. For the SRM, a stochastic reward function is used, where $y \overset{l}{\underset{U}{\rightarrow}} y'$ indicates that $y$ transitions to $y\prime$ on high-level label $l$, receiving reward sampled from the uniform distribution $U$.
    }
    \label{fig:prm}
\end{figure}

While traditional RMs have been proven effective in RL, several extensions and variants of RMs have been proposed recently to enhance their flexibility, scalability, and applicability to more complex tasks.
Dohmen et al.~\cite{dohmen2022inferring} propose the \textit{Probabilistic Reward Machine} (PRM), which uses probabilistic transition functions to capture the semantics of stochastic non-Markovian reward signals.
However, they only define stochastic rewards over a finite set of values, which restricts its ability to be generalized to broader scenarios. Instead,
Corazza et al.~\cite{corazza2022reinforcement} propose the \textit{Stochastic Reward Machine} (SRM),
which integrates stochastic reward functions to enhance the origin RM. 
Fig.~\ref{fig:prm} illustrates the major difference between the SRM and PRM.
Another key challenge of RMs is their limited scalability with increasing states. Furelos-Blanco et al.~\cite{furelos2023hierarchies} address this by introducing the \textit{Hierarchies of Reward Machines}, where each layer functions as a complete RM. They treats each RM call as an independently solvable subtask, facilitating learning at multiple levels of abstraction.
Furthermore, to equip RMs with the capabilities of solving inherently numeric tasks, Levina et al.~\cite{levina2024numeric} introduce the \textit{Numeric Reward Machine}, which incorporates numeric features into RMs.

In addition to the aforementioned methods, several studies further advance this field by utilizing RMs within regret minimization frameworks~\cite{bourel2023exploration}, enhancing RMs with high-level causal knowledge~\cite{corazza2023expediting}, 
using RMs to refine datasets for offline RL~\cite{sun2023less},
and integrating RMs into self-paced RL paradigms~\cite{koprulu2023reward}.
However, these aforementioned works share a common limitation, as the inherent semantics of RMs are based on regular grammar, which can only support tasks specified by regular LTL. Hence, finding ways to expand RMs to support tasks specified by full LTL continues to be a significant challenge.

\newcommand{\AM}{\makecell{Action\\ Model}}
\newcommand{\NMR}{Non-Markovian Reward} 
\newcommand{\MA}{MARL}
\newcommand{\cooperative}{Cooperative}
\newcommand{\adversarial}{Adversarial}
\newcommand{\QL}{Q-Learning}
\newcommand{\RL}{R-Learning}
\newcommand{\HRL}{HRL}
\newcommand{\IQL}{Independent Q-Learning}
\newcommand{\state}{S.}
\newcommand{\transition}{T.}
\newcommand{\reward}{R.}
\newcommand{\agent}{Ag.}
\newcommand{\option}{Op.}
\newcommand{\action}{Act.}
\newcommand{\office}{OfficeWorld}
\newcommand{\minecraft}{MineCraft}
\newcommand{\water}{WaterWorld}

\iftrue
\begin{table*}[htbp]
    \caption{KRR for \textit{Efficiency} of RL}
    \scriptsize
    \centering
    \renewcommand\arraystretch{1.3}
    \begin{tabular}{c|c|c|c|c|c}
    \hline\hline
       & KRR Techniques & Core Problems/Keywords & Base Algorithms & Reference & Experiment Domain   \\
    \hline
    \multirow{38}{*}{\makecell{Task \\ Representation}} 
    & \multirow{17}{*}{Reward Machines} & \NMR & \QL &  Icarte et al.\cite{icarte2018using,icarte2022reward} & \makecell{\office, \\ \minecraft, \water} \\\cline{3-6} 
    & & \MA & \IQL & Neary et al.\cite{neary2020reward} & Button, Rendezvous \\\cline{3-6}
    & & \MA  & \makecell{\QL,\\DQN} & Hu et al.\cite{hu2021decentralized} & \makecell{\office, \\ \minecraft, \water} \\\cline{3-6}
    & & \MA  & \QL & Zheng et al.\cite{zheng2024multi} & \makecell{Navigation, \\ \minecraft, Pass} \\\cline{3-6}
    & & Non-Predefined RM & \QL & Icarte et al.\cite{icarte2019learning, icarte2023learning} & Cookie, Symbol, 2-Keys \\\cline{3-6}
    & & Non-Predefined RM & \QL & Xu et al.\cite{xu2020joint} & \makecell{Vehicle, \\ \office, \minecraft} \\\cline{3-6}
    & & Non-Predefined RM & \QL & Furelos-Blanco et al.\cite{furelos2020induction} & \office\\\cline{3-6}
    & & Non-Predefined RM, MARL & \IQL & \makecell{Varricchione et al.\cite{varricchione2023synthesising},\\Ardon et al.\cite{ardon2023learning}} & Button, Rendezvous \\\cline{3-6}
    & & \makecell{Non-Predefined RM,\\Noisy Proposition} & PPO & \makecell{Hatanaka et al.\cite{hatanaka2023reinforcement}\\Li et al.\cite{li2024reward}} & \makecell{GridWorld, Robot Simulation} \\\cline{3-6}
    & & Non-Markovian Stochastic Reward & \QL &   Dohmen et al.\cite{dohmen2022inferring} &  \office \\\cline{3-6}
    & & Non-Markovian Stochastic Reward & \QL & Corazza et al.\cite{corazza2022reinforcement} &   Mining, Harvest\\\cline{3-6}
    & & Scalability & DQN & Furelos-Blanco et al.\cite{furelos2023hierarchies} & CraftWorld, WaterWorld \\\cline{3-6}
    & & Numeric Tasks & \QL & Levina et al.\cite{levina2024numeric} & MineCraft \\\cline{3-6}
    & & Low Regret & \QL & Bourel et al.\cite{bourel2023exploration}  & RiverSwim, Gridworld \\\cline{3-6}
    & & Casual Knowledge & \QL & Corazza et al.\cite{corazza2023expediting}  & Two-Doors, Coffee vs. Soda \\\cline{3-6}
    & & Offline RL & \QL & Sun et al.\cite{sun2023less}  & Kitchen, Adroit  \\\cline{3-6}
    & & Self-Paced RL & \QL & Koprulu et al.\cite{koprulu2023reward} & \makecell{Two-Door,\\ Swimmer-v3, HalfCheetah-v3} \\\cline{2-6}
    & \multirow{14}{*}{Temporal Logics} & \NMR & \makecell{Relative Entropy\\Policy Search} & Li et al.\cite{li2017reinforcement} & \makecell{Toast Placing,\\Simulated 2D Manipulation }\\\cline{3-6}
    & & \NMR & \makecell{Augmented\\Random Search} & Jothimurugan et al.\cite{jothimurugan2019composable}& \makecell{Cartpole, \\Robotic Motion Planning} \\\cline{3-6}
    & & High-Level Planning & \makecell{Augmented\\Random Search} & Jothimurugan et al.\cite{jothimurugan2021compositional} & Rooms, Fetch \\\cline{3-6}
    & & \NMR  & \QL & Camacho et al.\cite{camacho2019ltl} & \makecell{\office, \\ \minecraft, \water}\\\cline{3-6}
    & & Continuous State and Action Space & DDPG & Yuan et al.\cite{yuan2019modular} & \makecell{Melas Chasma,\\ Victoria Crater} \\\cline{3-6}
    & & Global Optimality & \makecell{\QL,\\PPO} & Voloshin et al.\cite{voloshin2023eventual,voloshin2022policy} & \makecell{ Minecraft, Pacman,\\ Flatworld, Carlo} \\\cline{3-6}
    & & MARL & \makecell{Alternating Direction\\ Method of\\ Multipliers} & Cubuktepe et al.\cite{cubuktepe2020policy} & Crop Fields, Urban Security  \\\cline{3-6}
    & & MARL  & MINLP Solvers & Djeumou et al.\cite{djeumou2020probabilistic}  & \makecell{GridWorld, \\ ROS-Gazebo Simulation} \\\cline{3-6}
    & & MARL & \makecell{Minimax Q-Learning,\\Minimax DQN} &  Muniraj et al.\cite{muniraj2018enforcing} & GridWorld \\\cline{3-6}
    & & MARL & \makecell{Independent\\Q-Learning} & Leon et al.\cite{leon2020extended} & \minecraft \\\cline{3-6}
    & & Underspecified LTL & Actor-Critic Methods & Den Hengst et al.\cite{den2022reinforcement} & \minecraft, Maze \\\cline{1-6} 
    \multirow{9}{*}{\makecell{Symbolic\\Planning}} 
    & \multirow{4}{*}{\makecell{Answer Set\\Programming}} & Safety, Exploration & \QL & Leonetti et al.\cite{leonetti2016synthesis} & \makecell{GridWord,\\ Robot Simulation, Service Robot}\\\cline{3-6}
    & & Non-Stationary Domains & \QL & Ferreira et al.\cite{ferreira2018method} & GridWorld  \\\cline{3-6} 
    & & HRL & PPO & Mitchener et al.\cite{mitchener2022detect} & Animal‑AI \\\cline{3-6}
    & &  Relational RL & \QL, SARSA & Nickles et al.~\cite{nickles2012integrating} & Blocks World \\\cline{2-6}
    & \multirow{5}{*}{Action Languages} & HRL & \RL & Yang et al.\cite{yang2018peorl} & Taxi, GridWorld \\\cline{3-6}
    & & HRL & \makecell{\RL,\\DQN} & Lyu et al.\cite{lyu2019sdrl} & Taxi, Montezuma's Revenges \\\cline{3-6}
    & & HRL & \QL & Illanes et al.\cite{illanes2020symbolic} &  \office, \minecraft \\\cline{3-6}
    & & HRL & \makecell{\QL,\\DQN} &  Kokel\cite{kokel2021reprel, kokel2022hybrid} &\makecell{MineCraft, OfficeWorld,\\ TaxiWorld, BoxWorld} \\\cline{2-6}
    \hline
    \hline
    \end{tabular}
    \label{tab:efficiency}
\end{table*}
\fi

\begin{figure*}[!t]
    \centering
    \subfloat[]{
        \includegraphics[height=0.18\linewidth]{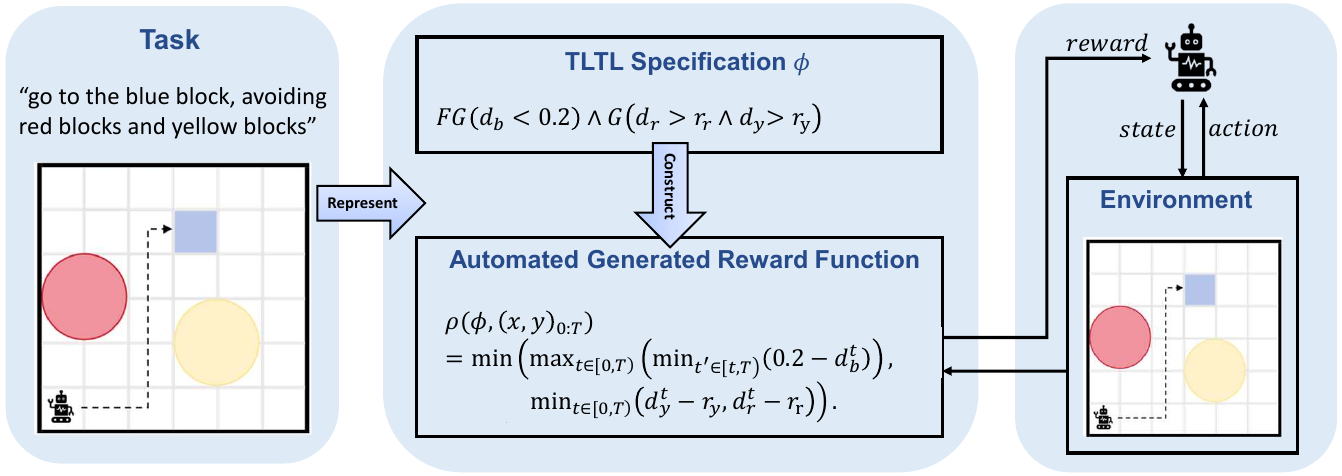}
        \label{fig:tltl}
    }
    \hfil
    \subfloat[]{
        \includegraphics[height=0.18\linewidth]{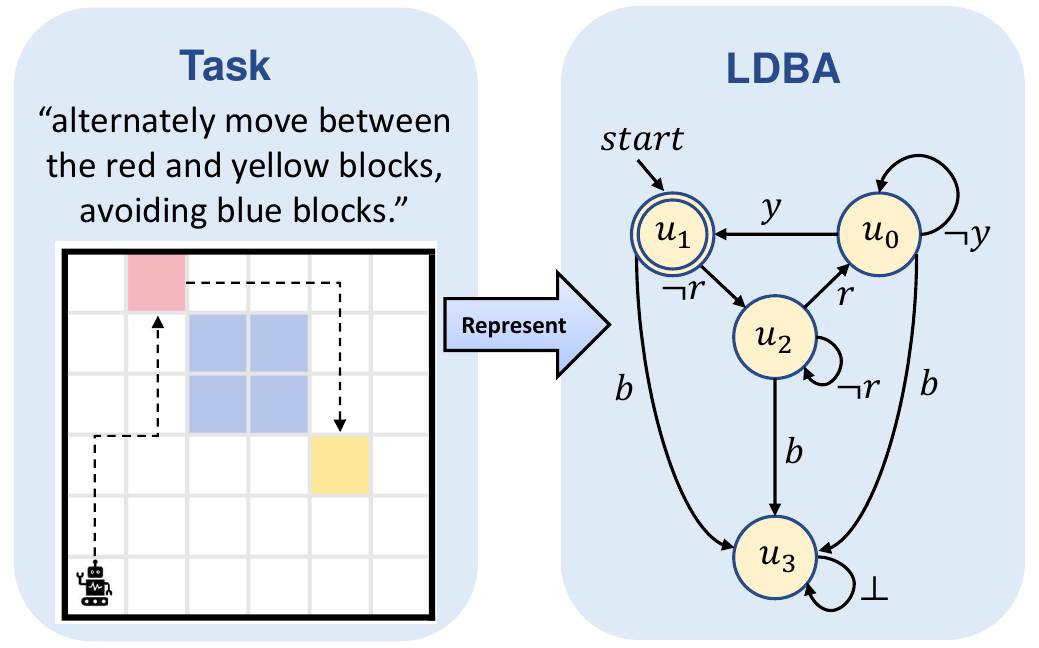}
        \label{fig:ldba_example}
    }
    \hfil
    \subfloat[]{
       \includegraphics[height=0.18\linewidth]{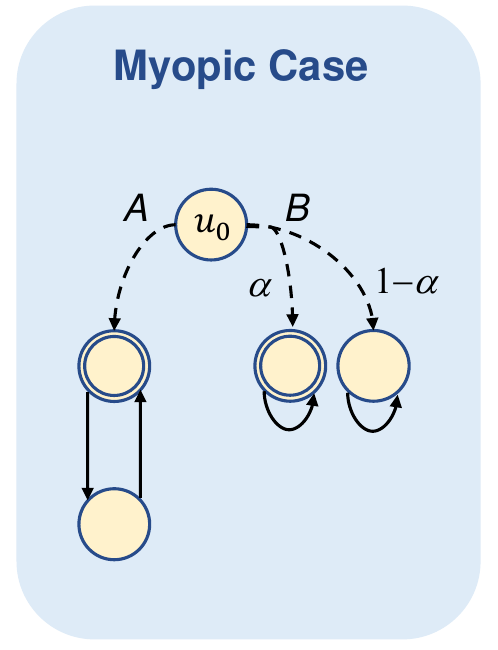}
        \label{fig:ldba_myopic}
    }
    
    \caption{An illustration of 
    (a) automatic reward function generation based on TLTL;
    (b) converting a task specified in $\omega$-regular LTL into a LDBA; and
    (c) a possible myopic case that may arise when utilizing LDBA (adapted from~\cite{voloshin2023eventual}).
    In (a), the task is defined by TLTL with quantitative semantics, enabling the automatic generation of a reward function that guides the agent.
  In (b), the task is achieved when the accepting states in the LDBA are accessed infinitely, with the only accepting state being $u_1$.
  In (c), an agent starts in state $u_0$ and only has two actions $A$ and $B$. Taking action $A$ transitions directly to an accepting state, from which the agent visits the accepting state every two steps. In contrast, action $B$ transitions to an accepting state with probability $\alpha$ and to a sink state with probability $1-\alpha$. However, when $\alpha$ exceeds a certain threshold, the agent may take the risk of choosing action $B$, leading to myopic behavior.
    }
    \label{fig:temporal_logic_algorithms}
\end{figure*}

\textbf{Temporal Logics.}
The normal RL problems modeled as traditional MDPs require to assign rewards according to a function of the last state and action. This formulation is often limited in domains where the rewards are not necessarily Markovian, but depend on an extended state space~\cite{toro2018teaching}. 
\textit{Non-Markovian} rewards can be specified in \textit{Temporal Logics} (TL) on finite traces such as LTL$_f$/LDL$_f$~\cite{brafman2018ltlf}, with the great advantage of a higher abstraction and succinctness.
There has been a large volume of work using specifications based on TL for specifying RL tasks~\cite{bozkurt2020control,de2019foundations,aksaray2016q,hasanbeig2018logically,hasanbeig2019reinforcement,littman2017environment,jiang2021temporal}. These approaches typically generate a (usually sparse) reward function from a given specification which is then used by an off-the-shelf RL algorithm to learn a policy.
In particular, Li et al.~\cite{li2017reinforcement} propose a variant of LTL called \textit{Truncated LTL} (TLTL) to specify tasks for robots. 
As shown in Fig.~\ref{fig:tltl}, by assigning quantitative semantics to TLTL formulas, logical formulas are converted into real-valued reward functions.
Jothimurugan et al.~\cite{jothimurugan2019composable} propose a specification language called SPECTRL that allows users to encode complex tasks involving sequences, disjunctions, and conjunctions of subtasks. 
Based on SPECTRL, Jothimurugan et al.~\cite{jothimurugan2021compositional} further propose a novel algorithm which breaks down a complex control problem into a high-level planning task and a series of low-level control tasks.
Camacho et al.~\cite{camacho2019ltl} show that RMs can be derived from specifications in various formal languages, including LTL and other regular languages.
Similarly, Yuan et al.~\cite{yuan2019modular} convert the LTL property to an automaton in order to synchronise the high-level LTL guide with RL.
Unlike most previous works, which focus on finite-state and finite-action MDPs, this study proposes a modular DDPG architecture for generating neural policies.

Most of the studies mentioned above utilize a key property of LTL, which is its equivalence to automata representations, such as the \textit{Limit Deterministic Büchi Automaton} (LDBA)~\cite{sickert2016limit}, as shown in Fig.~\ref{fig:ldba_example}.
There are several studies that leverage automata properties to facilitate the learning process.
For a task specified by LTL, Voloshin et al.~\cite{voloshin2022policy} propose a method that reduces a policy optimization problem into an automaton reachability problem.
To tackle myopic cases that may arise when using LDBA, as illustrated in Fig.~\ref{fig:ldba_myopic},
Voloshin et al.~\cite{voloshin2023eventual} reformulates the RL problem by maximizing the probability of satisfying LTL specifications without penalizing the time taken to achieve them. 
They develop a new experience replay technique, leveraging the structure of the LDBA to generate multiple off-policy trajectories.
However, these methods are not feasible without prior knowledge of the exact probability transition functions in the MDP~\cite{le2024reinforcement}, and are limited to discrete state/action spaces. 

\iftrue
Recently, researches have expanded to explore the use of temporal logic specifications in MARL. 
These works utilize the task structures revealed by temporal logic to decompose complex multi-agent tasks into simpler subtasks, thereby improving the learning efficiency and stability of RL agents.
In specific, Cubuktepe et al. \cite{cubuktepe2020policy} investigate the synthesis of policies for multi-agent systems to implement spatial-temporal tasks specified with  \emph{Graph Temporal Logic} (GTL) specifications.
Djeumou et al.~\cite{djeumou2020probabilistic} study a setting in which the agents move along the nodes of a graph, and the task specifications are expressed in GTL. 
The proposed algorithm is based on synthesizing a time-varying Markov matrix through the solution of an optimization problem.
Muniraj et al.~\cite{muniraj2018enforcing} investigate the problem of learning policies in an adversarial environment, where defensive agents strive to satisfy STL specifications against the adversarial agents' best responses. 
However, the proposed approach is applicable only in adversarial scenarios and is restricted to a specific type of temporal logic. Instead, Leon et al.~\cite{leon2020extended} formally define Extended Markov Games as a general mathematical model that allows multiple RL agents to learn various non-Markovian specifications. 
These works are designed to tackle finite-horizon tasks, which limits their applicability. In contrast, some studies have begun to explore addressing infinite-horizon tasks in a multi-agent setting\cite{hammond2021multi,zhu2024decomposing,terashima2024reward}.
\fi


The above studies mostly rely on the assumption that the high-level knowledge (i.e., \textit{temporal logic specifications}) are given, while in reality, acquiring such detailed instructions can be quite challenging, with many being even implicit and needing to be inferred from data.
Den Hengst et al.~\cite{den2022reinforcement} introduce \textit{Option Machines}, which enhance sample efficiency by optimizing rewards under underspecified LTL instructions. 
Furthermore, there are already some studies that infer temporal logic specifications from data, using \textit{Bayesain inference}~\cite{ShahKSL18bayesian,KimMSAS19bayesian}, \textit{hidden Markov models} \cite{abate2023learning} or \textit{direct search} techniques~\cite{CamachoM19learning}. However, how to conduct this high-level model learning in RL, especially in noisy and partial settings~\cite{LuoLDWPZ22Bridging}, is still a challenging problem in this area.

\subsection{Symbolic Planning}




RL and symbolic planning have both been used to build intelligent agents that conduct sequential decision making in uncertain environments. Those methods, however, have their own focuses: RL focuses on interactions with the world, but often requires an unfeasibly large amount of experience, while planning focuses on utilizing prior knowledge, but relies strongly on the accuracy of the model.
There has been an increasing interest in combining high-level symbolic planning with RL in order to further improve the learning performance. 

\textbf{Answer Set Programming.}
Recent studies have shown how the integration of \emph{Answer Set Programming} (ASP) with RL can enhance decision-making efficiency.
Leonetti et al.~\cite{leonetti2016synthesis} present a method that takes advantage of planning through ASP to constrain the behavior of the agent by discarding clearly suboptimal plans and creating a partial policy of reasonable actions. 
Unlike~\cite{leonetti2016synthesis} that requires defined sets of states and actions, along with the transition function,  Ferreira et al.~\cite{ferreira2018method} propose an algorithm that uses ASP rules to constrain the action space, allowing for real-time updates to decision-making policies in an unknown environment. 
Some other works~\cite{nickles2012integrating,mitchener2022detect} propose a method that uses ASP to integrate a formal calculus for reasoning about actions into relational RL agents, and a neuro-symbolic HRL framework that combines computer vision for detection, a symbolic meta-policy using ASP and ILP for efficient learning and reasoning and DRL-based options for low-level decision-making. 
However, most of the work in this area is still focused on simple cases involving MDPs with discrete state and action space, typically using basic RL algorithms such as Q-Learning. 
How to combine ASP techniques with more advanced RL algorithms in more intricate scenarios, like multi-agent settings, is still a remaining challenge.

\begin{figure*}[!t]
    \centering
    \subfloat[]{
        \includegraphics[width=0.27\linewidth]{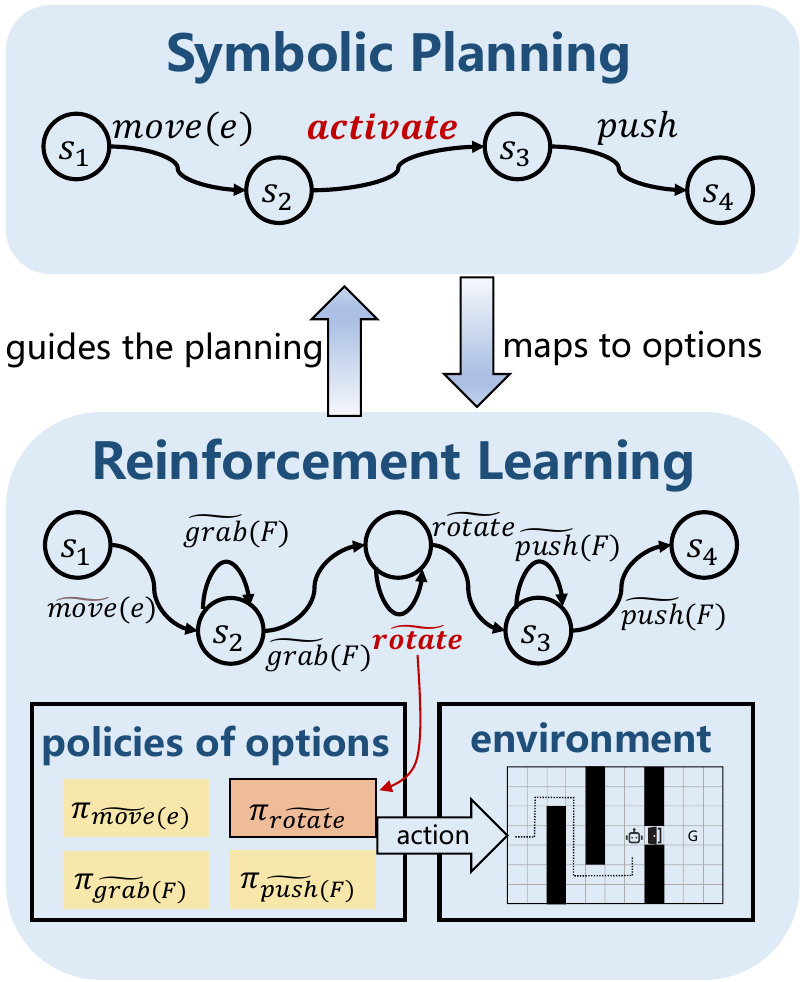}
        \label{fig:symbolicplanning}
    }
    \hfil
    \subfloat[]{
        \includegraphics[width=0.33\linewidth]{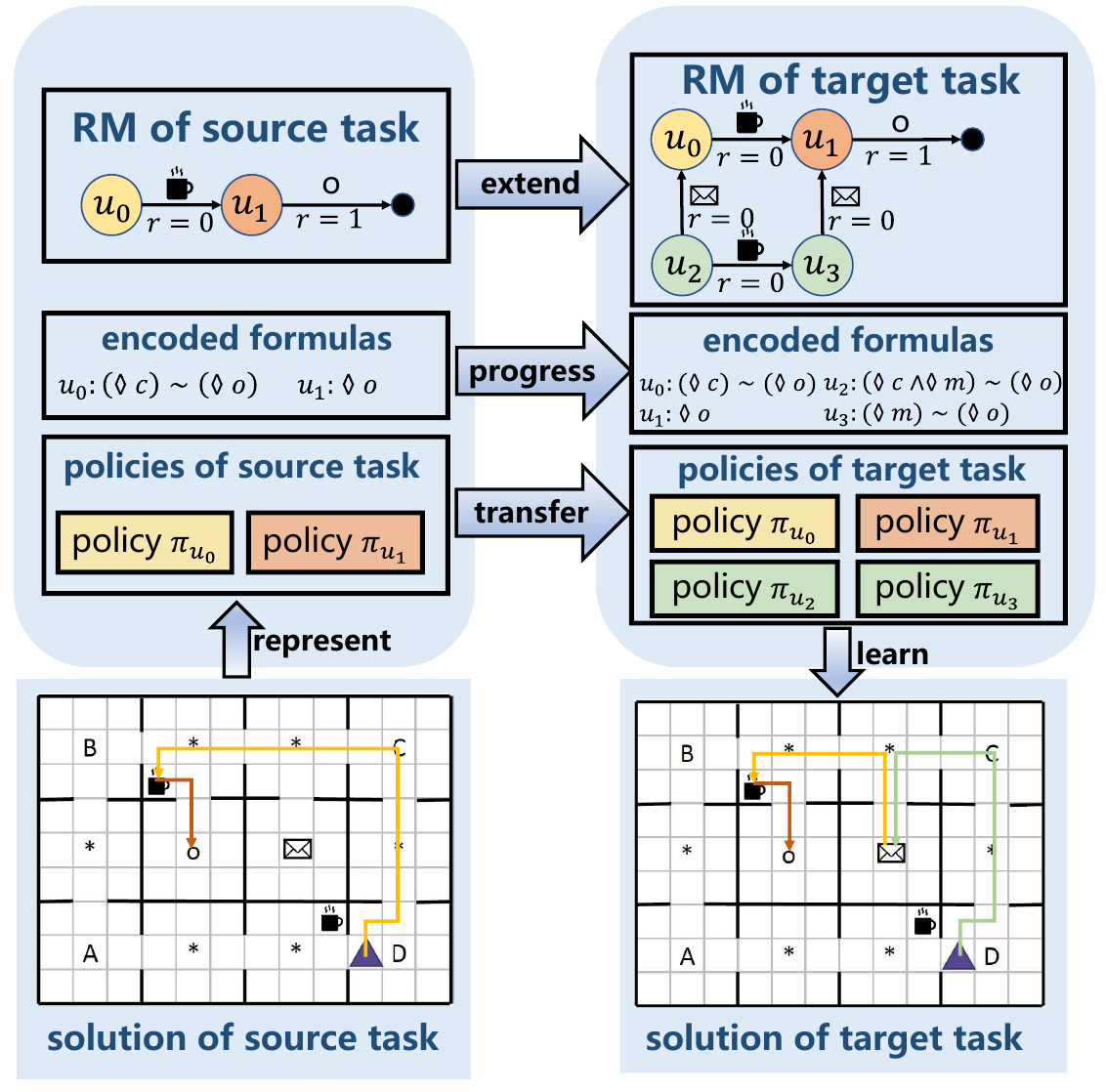}
        \label{fig:LSRM}
    }
    \hfil
    \subfloat[]{
        \includegraphics[width=0.3\linewidth]{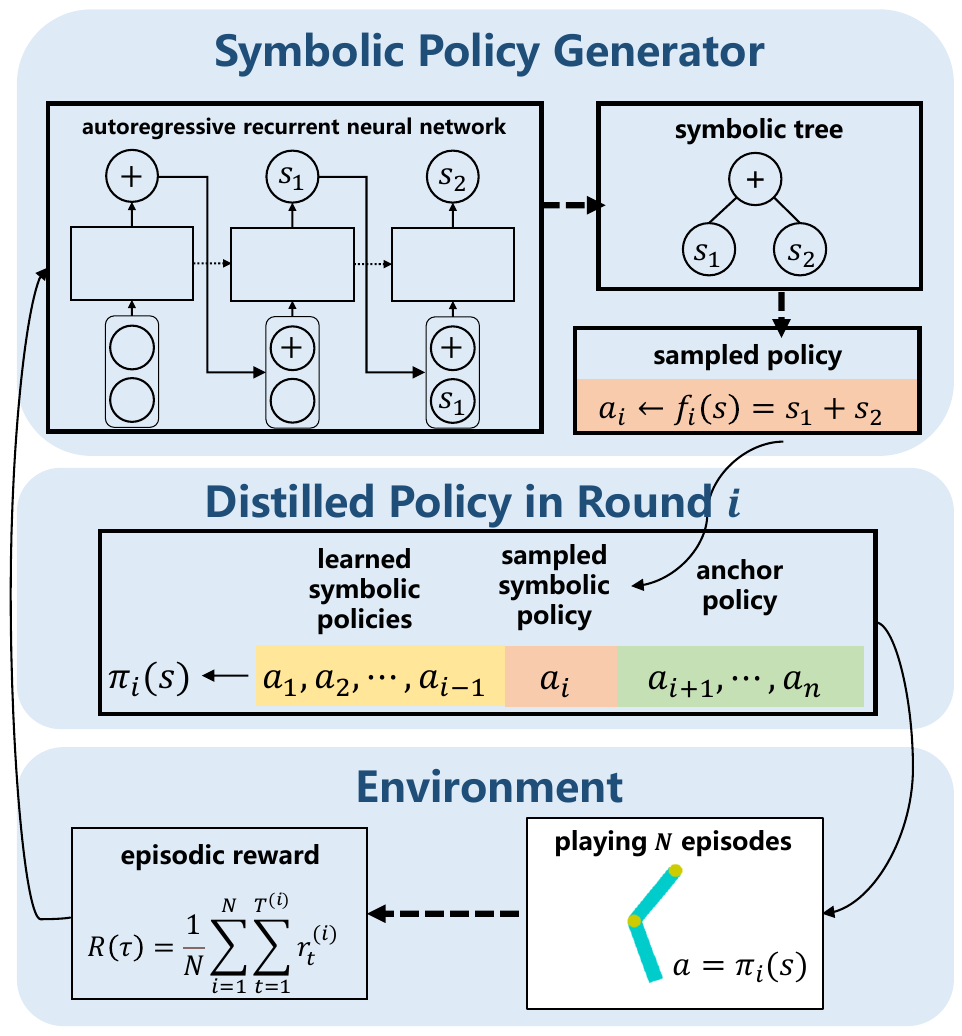}
        \label{fig:symbolic_policy}
    }
    
    \caption{An illustration of 
    (a) symbolic planning with RL;
    (b) transferring knowledge from a source task to a target task formulated as RMs; and
    (c) learning deep symbolic policy (adapted from~\cite{landajuela2021discovering}).
    In (a), an agent learns to move to the door, open the door, then reach the goal. Symbolic planning gives the symbolic solution of the current action model: $move(e), activate$, then $push$. These symbolic actions are mapped to options in RL: $\widetilde{move}(e), \widetilde{grab}(F)$, etc. The optimal option $\widetilde{rotate}$ are then chosen by the high-level policy, and its corresponding low-level policy $\pi_{\widetilde{rotate}}$ chooses an action to be executed in the environment. 
    (b) gives an illustration of transferring knowledge from a source task \textit{``get coffee then go to office''} to a target task \textit{``get mail and coffee then go to office''}. The source task is represented as an RM with 2 states, both of which are encoded with SLTL formulas. For example, the state $u_0$ is encoded with formula $(\Diamond c)\sim (\Diamond o)$, which means \textit{``get coffee first, then go to office''}. The SLTL formula of the target task is progressed to generate 2 new formulas, which serves as new states $u_2$ and $u_3$ of the extended RM. Two policies of the source task are also transferred to the target task by directly copying Q-functions of RM states that share with the same encoded formulas, and then composing the Q-functions according to the constructions of the newly generated formulas. 
    In (c), a symbolic policy is structured as a tree and generated by an autoregressive RNN. The generated policy is evaluated in a specific environment, and the resulting trajectories are collected to update the RNN.
    }
    \label{fig:overall}
\end{figure*}

\textbf{Action Languages.}
Some works have integrated action languages with symbolic planning with RL, where symbolic planning functions as a high-level policy and low-level policies are implemented by RL algorithms.
Yang et al.~\cite{yang2018peorl} present a unified framework PEORL that integrates symbolic planning with HRL to cope with decision-making in a dynamic environment with uncertainties. In PEORL, symbolic plans are used to guide the agent's task execution and learning, and the learned experience is fed back to symbolic knowledge to improve planning. 
During training, quality is assessed based on returns, while a goal guides subsequent iterations to generate higher-quality plans.
Later, Lyu et al.~\cite{lyu2019sdrl} further generalize the goal of the PEORL framework~\cite{yang2018peorl} and enhance it by integrating DRL to handle both high-dimensional sensory inputs and symbolic planning. 
The proposed SDRL method includes a planner, a controller and a meta-controller, as demonstrated in Fig.~\ref{fig:symbolicplanning}, which takes charge of subtask scheduling, data-driven subtask learning, and subtask evaluation, respectively. 
The three components cross-fertilize each other and eventually converge to an optimal symbolic plan and subpolicies, bringing together the advantages of long-term planning and end-to-end RL.

The aforementioned works primarily focus on generating meaningful intermediate goals to accelerate the learning of policies for tasks with unknown reward functions. To allow users to define tasks as they wish and reduce the burden of reward design,
Illanes et al.~\cite{illanes2020symbolic} explore the use of high-level symbolic PDDL action models as a framework for defining final-state goal tasks and automatically producing their corresponding reward functions.
Building on the Taskable RL framework~\cite{illanes2020symbolic}, Kokel et al.~\cite{kokel2021reprel} propose RePReL by incorporating two significant extensions: generalizing the Taskable RL framework to the relational MDP setting and introducing a novel approach for defining task-specific state abstractions, which are crucial for improving transferability and generalization. 
Later, Kokel et al.~\cite{kokel2022hybrid} extend the RePReL framework to the DRL setting. 


The above studies, however, all assume that the high-level planning models should be provided by domain experts. To mitigate this limitation, Guan et al.\cite{guan2022leveraging} utilize incorrect and incomplete symbolic models to address long-horizon, goal-directed tasks with sparse rewards. Futhermore,
instead of assuming planning model provided as input, MuZero~\cite{DBLP:journals/nature/SchrittwieserAH20} learns an environment model, which consists of an encoder network for capturing abstract states, a dynamics network for capturing the transition of abstract states, and a reward network. It exploits the learnt model to do MCTS planning for repeatedly improving the policy. Similarly, EfficientImitate~\cite{DBLP:conf/nips/Yin22} learns an environment model by extending \textit{LAdversarial Imitation Learning} into the MCTS planning-based RL. A recent study~\cite{jin2022creativity}  utilizes the collected trajectories from the lower level of RL to learn PDDL action models and symbolic options, 
while Wu et al.~\cite{wu2022models,wu2022Plan} direct estimate the action models in order to generate more samples for efficient policy learning.
Additionally, there is an increasingly body of research dedicated to leveraging \textit{Large Language Models} (LLMs) for the generation of PDDL models~\cite{mahdavi2024leveraging,guan2023leveraging,smirnov2024generating}.

\section{KRR for \textit{Generalization} of RL}
\label{sec:four}

Much current RL research evaluates the learned policy in exactly the same environment where the policy is trained on. This assumption contradicts real-world situations when the training and testing environments might be different. 
Generalisation in RL~\cite{kirk2021survey}  aims to propose RL algorithms that generalise well to unseen situations at deployment time by avoiding overfitting to the training environments (i.e., the \textit{one-shot transfer learning} problem~\cite{kirk2021survey})  or transferring knowledge learned from previous tasks to solve new but related tasks quickly (the \textit{lifelong or continual RL} problem~\cite{khetarpal2020towards}).  KRR techniques have been widely applied in both settings by directly transferring a logically-specified policy to a new task, or solving more complex tasks that are expressed as logical compositions of previously learned sub-tasks.
Table \ref{tab:generalization} provides a summary of how KRR techniques can be employed to enhance the generalization capabilities of RL agents.

\subsection{Transfer Learning}
The sampling efficiency and performance of RL can be improved if some high-level knowledge can be transferred from a source task to a target task once these tasks are logically similar. Some studies resort to the powerful expression capabilities of formal logic to facilitate transfer learning in RL, either using Markov Logic Networks, Temporal Logics, or Reward Machines. 

\textbf{Markov Logic Networks (MLNs).} 
There have been some works that utilize the capabilities of MLNs,  a statistical relational learning model that combines Markov networks with FOL, to enhance the transferability of RL.
Mihalkova et al.~\cite{mihalkova2007mapping} propose a transfer learning system that autonomously maps a source task expressed as an MLN to the target domain and then revises its logic structure to further improve the learning performance.
However, in cases when only partial data about the target domain is available, this method is likely to perform suboptimally. 
To address this limitation, the author~\cite{mihalkova2008transfer} further propose a new algorithm which evaluates possible source-to-target predicate correspondences based on short-range clauses, in order to transfer the knowledge captured in long-range clauses.
Torrey et al.~\cite{torrey2010policy} further propose value function and policy based transfer learning algorithms via MLN and evaluate them in a complex RoboCup simulated-soccer domain. 
However, a significant limitation in this area is the scalability of MLNs, as the computational cost of learning MLN structures and weights remains challenging, particularly in complex domains~\cite{mansour2017recent}. 


\textbf{Temporal Logics (TLs).}
Other studies have also explored the combination of RL and TLs to facilitate transfer beyond the training set by leveraging the revealed commonalities between source and target tasks.
In specific, for each task specified by LTL, Li et al.~\cite{li2017automata} transform it to a corresponding FSA, 
which represents a hierarchical policy with low-level controllers, and policy transfer can be achieved by re-modulating low-level controllers of different FSAs, as shown in Fig.~\ref{fig:fsa}.
However, the transfer method only considers new tasks that are conjunction ($\land$) combinations of existing source tasks.
León et al.~\cite{leon2020systematic} propose a neuro-symbolic architecture that combines DRL with TLs to achieve systematic out-of-distribution generalisation in formally specified tasks, 
including combinations of training tasks using negation ($\neg$) and disjunction ($\lor$) operators. 
This subtask-based approach may lead to suboptimal outcomes or even violate the task specifications, as individual subtasks might not align perfectly with the overall goals of the main task.
To address this issue, Xu et al.~\cite{xu2024generalization2} propose a zero-shot transfer method that incorporates a novel implicit planner to select subsequent subtasks and estimate the returns for completing the remaining tasks.
All these works, however, still assume that the interpretations of high-level propositions in the environment are fixed.

\textbf{Reward Machines (RMs).}
RMs can also be employed for transfer learning, as similar states or transitions in RMs represent similar components across different tasks.
Liu et al.~\cite{liu2024skill} introduce a zero-shot transfer algorithm that combines task-agnostic policies learned during training to safely satisfy a wide variety of novel LTL task specifications.
Task-agnostic policies are derived from state-centric policies that are trained using the method in~\cite{toro2018teaching}.
Azran et al.~\cite{azran2024contextual} propose a few-shot transfer method that leverages knowledge from previously encountered propositions and transitions. Rather than concentrating on the states of RMs, they utilize similar transitions within RMs to facilitate the transfer process.
The main limitation of this method is that its effectiveness in facilitating transfer heavily relies on the resolution of the generated RMs.
Moreover, a sub-policy from one RM may not be the optimal solution for a sub-policy of another RM, which could affect its transfer performance.

While significant progress has been made, efficient transfer learning between tasks with logic expressions and temporal constraints is still a complicated issue due to the difficulties in formally defining and explicitly quantifying logical similarities between tasks and automatically specifying what knowledge to be transferred based on these similarities. 
A notable work~\cite{xu2019transfer} tries to provide a solution to this problem by computing the similarity between temporal tasks through a notion of \textit{logical transferability}, and developing a transfer learning approach based on this quantification.
They first propose an inference technique to extract \textit{Metric Interval Temporal Logic} (MITL) formulas in sequential disjunctive normal form from labeled trajectories collected in RL of the two tasks. If logical transferability is identified through this inference, they construct a timed automaton for each sequential conjunctive subformula of the inferred MITL formulas from both tasks.  Mappings between corresponding components of the automata are then established to transfer Q-functions.

\iftrue
\begin{table*}[htbp]
    \caption{KRR for \textit{Generalization} of RL}
    \scriptsize
    \centering
    \renewcommand\arraystretch{1.3}
    \begin{tabular}{c|c|c|c|c|c}
    \hline\hline
       & KRR Techniques & Core Problems/Keywords & Base Algorithms & Reference & Experiment Domain   \\
    \hline
    \multirow{7}{*}{\makecell{Transfer \\ Learning}} & \multirow{2}{*}{\makecell{Markov\\Logic Networks}} & FOL Rules Transfer & Relational Pathfinding & Mihalkova et al.\cite{mihalkova2007mapping, mihalkova2008transfer} & IMDB, UW-CSE, WebKB\\ \cline{3-6}
    & & FOL Rules Transfer & Q-Learning & Torrey et al.\cite{torrey2010policy} & RoboCup Simulated-Soccer \\ \cline{2-6}
    & \multirow{4}{*}{Temporal Logics}  & Few-Shot Policy Transfer & Q-Learning & Li et al.\cite{li2017automata} & GridWorld, Kitchen\\ \cline{3-6}
    & & Zero-Shot Generalization & A2C & León et al.\cite{leon2020systematic} & MineCraft \\ \cline{3-6}
    & & Zero-Shot Generalization & PPO & Xu et al.\cite{xu2024generalization2} & LetterWorld, Walk, Zone\\ \cline{3-6}
    & & Few-Shot Policy Transfer & Q-Learning & Xu et al.\cite{xu2019transfer} & GridWorld\\ \cline{2-6}
    & \multirow{2}{*}{Reward Machines} & Zero-Shot Policy Transfer & DQN & Liu et al.\cite{liu2024skill} & MineCraft, Physical Robot \\ \cline{3-6}
    & & Few-Shot Policy Transfer & DQN & Azran et al.\cite{azran2024contextual} & GridWorld\\ \cline{2-6}
    \hline
    \multirow{14}{*}{\makecell{Lifelong \\ Learning}}  & \multirow{10}{*}{\makecell{Temporal Logics}}  & Generalization through DNN & A2C & Kuo et al. \cite{kuo2020encoding} & Symbol, MineCraft\\ \cline{3-6}
    & & Generalization through DNN & PPO & Vaezipoor et al.\cite{vaezipoor2021ltl2action} & LetterWorld, Zone\\ \cline{3-6}
    & & Generalization through DNN & A2C & León et al.\cite{leon2021nutshell} & MineCraft, MiniGrid\\ \cline{3-6}
    & & Value Function Composition & DQN & Van et al. \cite{van2019composing} & GridWorld\\ \cline{3-6}
    & & Value Function Composition & Q-Learning, DQN & Tasse et al.\cite{tasse2020boolean,tasse2020logical,tasse2021generalisation} & GridWorld \\ \cline{3-6}
    & & Value Function Composition & Q-Learning, DQN, TD3 & Tasse et al.\cite{tasseskill} & OfficeWorld, Moving Targets, Safety Gym \\ \cline{3-6}
    & & Off-Policy Learning & Q-Learning, DQN & Icarte et al. \cite{toro2018teaching} & MineCraft \\ \cline{3-6}
    & & Goal-Conditioned RL & PPO & Qiu et al.\cite{qiu2024instructing} & Zone, Ant-16Rooms, LetterWorld\\ \cline{3-6}
    & & Global Optimality & DQN, PPO & Xu et al.\cite{xu2024generalization} & LetterWorld, MineCraft, Safety Gym\\ \cline{3-6}
    & & Bottom-Up Composition & Q-Learning & Zheng et al.\cite{zheng2022lifelong} & OfficeWorld, MineCraft\\ \cline{3-6}
    & & Successor Feature & Q-Learning & Kuric et al.\cite{kuric2024planning} & Delivery, OfficeWorld\\ \cline{2-6}
    & \multirow{4}{*}{Natural Languages} & Grounding Knowledge & PPO & Hanjie et al. \cite{hanjie2021grounding} & Messenger, RTFM, BabyAI\\ \cline{3-6}
    & & Grounding Knowledge & PPO & Cao et al. \cite{cao2024exploring} & Messenger\\ \cline{3-6}
    & & Meta RL & \makecell{Maximum a Posteriori\\Policy Optimisation} & Bing et al. \cite{bing2023meta} & Meta-World \\ \cline{3-6}
    & & Auxiliary Task Generation & Q-Learning, DQN & Quartey et al. \cite{quartey2023exploiting} & HomeGrid \\ \cline{3-6}
    \hline
    \hline
    \end{tabular}
    \label{tab:generalization}
\end{table*}
\fi

\subsection{Continuous/Lifelong Learning}
Apart from the above zero-shot learning by transferring a single policy to improve learning on a new task, a more realistic and general setting for generalisation in RL is the continuous/lifelong learning problem when an agent is presented with a series of tasks sampled from some distribution or organized as a curriculum.


\textbf{Temporal Logics (TLs).}
There is currently a collection of research that leverages the generalization capabilities of neural networks to tackle tasks defined by TLs.
Kuo et al.~\cite{kuo2020encoding} propose to parse an LTL formula into a tree structure, with nodes representing operators or predicates. Each node is implemented using an RNN and interconnected through recurrent connections, and the tree node is used to output actions.
Vaezipoor et al.~\cite{vaezipoor2021ltl2action} employ an LTL module that uses graph neural networks to effectively represent the tree structure of LTL formulas.
Building upon~\cite{leon2020systematic}, León et al.~\cite{leon2021nutshell} introduce a novel neural-symbolic architecture that incorporates inductive bias to enhance generalization by linking task-agnostic representations of the current state to inferred latent goals. 
However, these approaches that use neural networks for generalization often demand a lot of training data, even in relatively simple environments. One possible solution to this challenge is to implement data augmentation techniques, as shown in~\cite{mumuni2022data}.

\begin{figure*}[!t]
    \centering
    \subfloat[]{
        \includegraphics[width=0.48\linewidth]{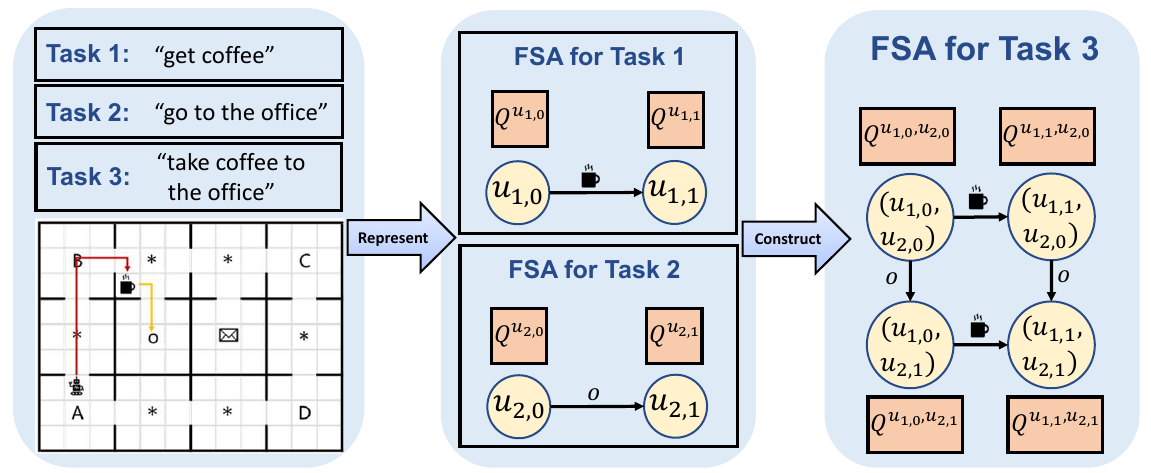}
        \label{fig:fsa}
    }
    \hfil
    \subfloat[]{
        \includegraphics[width=0.49\linewidth]{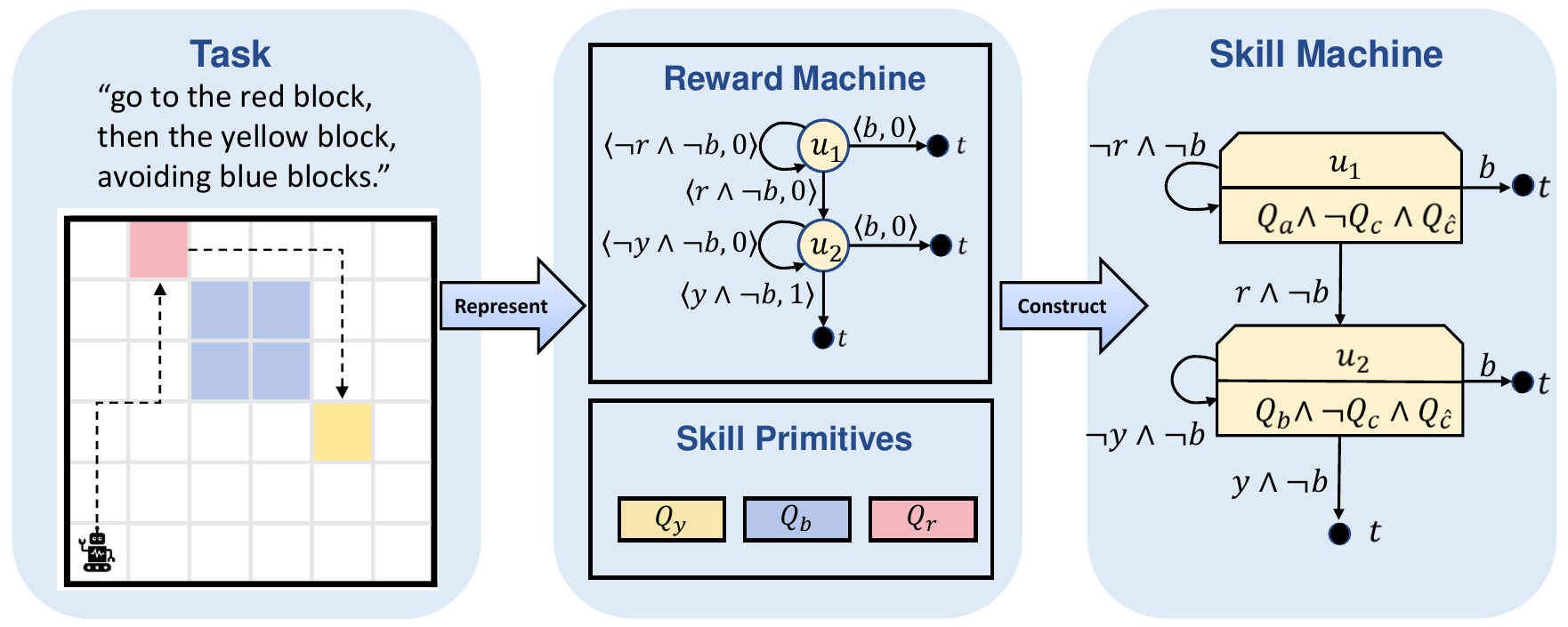}
        \label{fig:skill_machine}
    }
    
    \caption{An illustration of
    (a) knowledge transfer through composition of automata and value functions; and
    (b) construction process of a skill machine.
    In (a), consider that the agent has learned the policies for Task 1, \textit{``get coffee,''} and Task 2, \textit{``go to the office.''} Now, for a new Task 3, \textit{``take coffee to the office,''} an automaton can be constructed accordingly. For a certain state of the automaton $(u_{1,\alpha},u_{2,\beta})$, the corresponding value function $Q^{(u_{1,\alpha},u_{2,\beta})}$ can be initialized as the sum of the value functions from the source tasks, represented as $Q^{u_{1,\alpha}}+Q^{u_{2,\beta}}$. 
    In (b), to build a skill machine, it is essential to first create a corresponding RM structure based on the LTL specification, while also learning a set of skill primitives. Once this foundation is established, value composition is employed to initialize the polices of the skill machine.
    }
    \label{fig:automaton_algorithms}
\end{figure*}



Some recent studies focus on composing value functions as an alternative strategy for achieving generalization across new tasks specified by TLs.
Van Niekerk et al.~\cite{van2019composing} prove that value functions in  entropy-regularised RL can be optimally composed to solve new tasks that are the union of previous tasks.
While only considering the conjunction ($\land$) and disjunction ($\lor$) of tasks without formal definitions, Tasse et al.~\cite{tasse2020boolean} advance the field by demonstrating zero-shot logical composition of tasks without additional assumptions, additionally considering new tasks in terms of the negation ($\neg$) of base tasks. 
The authors~\cite{tasse2020logical,tasse2021generalisation} also demonstrate how such a framework can be used to tackle the lifelong learning problem
which provides guarantees on the generalisation of an agent's skills over an unknown task distribution. 
Later, Tasse et al.~\cite{tasseskill} present the \textit{Skill Machine} (SM) to improve the efficiency and combinatorial generalization performance of the RM by introducing value combination methods~\cite{tasse2020boolean} to achieve both spatial and temporal composition, as shown in Fig.~\ref{fig:skill_machine}.
While these value combination methods effectively leverage the characteristics of temporal logics, they only consider a subset of temporal logic operators. There is still potential to further explore the relationship between full LTL operators and value combinations.


There is ongoing exploration on how the formal properties of TLs and corresponding automata can be leveraged to enhance the generalization capabilities of RL agents. 
Icarte et al.~\cite{toro2018teaching} propose a novel algorithm called LPOPL, which leverages LTL progression to break down complex tasks into simpler, independently learnable components. Off-policy RL is employed to simultaneously update the value functions for all subtasks.
The algorithm proposed by Kuric et al.~\cite{kuric2024planning} integrates \textit{successor features} (SFs) with FSA to achieve zero-shot generalization. 
In cases where a single symbolic state maps to multiple MDP states, previous works often encounter suboptimal outcomes.
Xu et al.~\cite{xu2024generalization} propose a novel method that learns options based on future subgoal sequences to mitigate this issue. 
However, these algorithms~\cite{toro2018teaching,kuric2024planning,xu2024generalization} are restricted to regular temporal logic expressions, which only account for the finite behavior of the agent. 
Instead, Qiu et al.~\cite{qiu2024instructing} propose a method that uses Dijkstra’s algorithm to generate goal sequences from graph representations of tasks, enabling goal-conditioned RL agents to follow arbitrary $\omega$-regular LTL specifications without additional training.  
Their method may still suffer from suboptimality due to the myopic goal-conditioned value functions~\cite{yalcinkaya2024compositional}. 
Besides, LTL and the corresponding automata do not inherently account for uncertainty or stochastic behaviors, which can be critical in real-world applications.

Most existing work along this line requires that the source of subtasks should be specified a priori and the focus is on bottom-up composition effectiveness. To address this issue, Zheng et al.~\cite{zheng2022lifelong} propose a lifelong RL method that is able to learn and decompose logically complex tasks automatically through a lifelong memory. The authors first propose \textit{Sequential LTL} (SLTL), a supplemental formal language of LTL to enable more flexible specification of temporal tasks, and then use RMs to exploit structural reward functions for tasks encoded with high-level events. An automatic extension of RM for continuous learning over lifetime is then proposed to transfer knowledge from the learned Q-functions to the target tasks that are formulated as an SLTL formula.
As demonstrated in Fig.~\ref{fig:LSRM}, by storing and leveraging high-level knowledge in memory, the method is able to achieve systematic out-of-distribution generalization in tasks that follow the specifications of formal languages.

\textbf{Natural Languages.}
Natural language offers a rich and intuitive way to express user needs, as it doe not require knowledge of the semantics of common KRR techniques, such as TLs.
Specifically, for tasks described by text, Hanjie et al.~\cite{hanjie2021grounding} generalize their policies to new environments by employing a multi-modal entity-conditioned attention module and learning the latent grounding of entities and dynamics.
Cao et al.~\cite{cao2024exploring} further enhance the generalization capability of this framework by incorporating behavior prediction to account for the behavioral information in the text.
Bing et al.~\cite{bing2023meta} propose a meta-RL algorithm where the agent receives language descriptions of tasks in the training phase to understand its goals, and then uses this information to explore and solve new tasks in the adaptation phase.
In particular, in recent years, pre-trained \textit{Large Language Models} (LLMs) have emerged as a promising approach to enhance RL agents, owing to the inherent prior knowledge they possess and their exceptional reasoning capabilities. These attempts include using LLMs to enhance RL agents in reward generation~\cite{rocamonde2023vision}, value function initialization~\cite{yu2023b}, zero-shot generalization~\cite{zhao2023test} and more.
For instance, Quartey et al.~\cite{quartey2023exploiting} propose a framework to leverage LLMs to generate auxiliary tasks for target LTL tasks, enhancing the agent's capability for multi-task learning.
With the assistance of LLMs, RL agents are able to understand and extract knowledge about their environment more efficiently, including essential facts and rules, while KRR techniques can provide a rigorous framework for structuring this knowledge. Thus, how to combine LLMs with formal KRR techniques is a promising way to address complex RL problems~\cite{cao2024survey,rashidi2024survey}.

\section{KRR for \textit{Safety/Interpretability} of RL}
\label{sec:five}


Safety and interpretability are crucial in real-world RL applications, where agents may operate in uncertain and potentially dangerous environments, such as autonomous driving~\cite{kiran2021deep}, healthcare~\cite{yu2021reinforcement}, and industrial robotics~\cite{nian2020review}. KRR techniques can enhance the safety of RL systems by incorporating logical constraints and safety rules that ensure agents operate within predefined safe limits. Additionally, KRR improves interpretability by providing symbolic explanations for the agent's actions, enabling users to understand the decision-making process. This transparency is particularly vital in safety-critical domains, as it enables verification and fosters trust in the agent's behavior.
A summary of KRR techniques for improving the safety and interpretability of RL systems can be found in Table \ref{tab:safety}.

\iftrue
\begin{table*}[htbp]
    \caption{KRR for \textit{Safety/Interpretability} of RL}
    \scriptsize
    \centering
    \renewcommand\arraystretch{1.3}
    \begin{tabular}{c|c|c|c|c|c}
    \hline\hline
       & KRR Techniques & Core Problems/Keywords & Base Algorithms & Reference & Experiment Domain   \\
    \hline
    \multirow{19}{*}{\makecell{Interpretability}} & \multirow{4}{*}{\makecell{Interpretable\\Rewards}}  & Evolutionary Algorithms & SAC, Maximin DQN & Sheikh et al.\cite{sheikh2020learning} & Mujoco, Pygame, OpenAI-Gym\\ \cline{3-6}
    & & Inverse RL & PPO & Bougie et al.\cite{bougie2023interpretable} & Mujoco \\ \cline{3-6}
    & & Inverse RL & \makecell{PPO, Adversarially\\Guided Actor-Critic} & Zhou et al.\cite{zhou2022programmatic} & MiniGrid\\ \cline{2-6}
    & \multirow{5}{*}{\makecell{Symbolic\\Policies}}  & \makecell{Model-Based RL,\\Evolutionary Algorithms} & Monte Carlo Method & Hein et al.\cite{hein2018interpretable} & OpenAI-Gym\\ \cline{3-6}
    & & Model Distillation & DQN, PPO & Bastani et al.\cite{bastani2018verifiable} & OpenAI-Gym\\ \cline{3-6}
    & & Model Distillation & REINFORCE & Landajuela et al.\cite{landajuela2021discovering} & OpenAI-Gym \\ \cline{3-6}
    & & \makecell{Gradient-Based\\Symbolic Policy Learning} & Gradient Descent & Guo et al.\cite{guo2024efficient} & OpenAI-Gym\\ \cline{2-6}
    & \multirow{7}{*}{\makecell{Programmatic\\Polices}} & Model Distillation & DDPG, Duel-DDQN & Verma et al.\cite{verma2018programmatically} & TORCS, OpenAI-Gym \\ \cline{3-6}
    & & Model Distillation & DDPG & Verma et al.\cite{verma2019imitation} & TORCS, OpenAI-Gym \\ \cline{3-6}
    & & Relational RL & Vanilla Policy Gradient & Jiang et al.\cite{jiang2019neural} & Block Manipulation, Cliff Walking\\ \cline{3-6}
    & & \makecell{Programs with\\Cause-Effect Logic} & Vanilla Policy Gradient & Cao et al.\cite{cao2022galois} & MiniGrid \\ \cline{3-6}
    & & \makecell{Programs without Sketches} & Cross Entropy Method & Trivedi et al.\cite{trivedi2021learning} & Karel \\ \cline{3-6}
    & & \makecell{Programs without Sketches} & PPO, SAC & Liu et al.\cite{liu2023hierarchical}  & Karel \\ \cline{3-6}
    & & \makecell{Programs without Sketches} & PPO, Cross Entropy Method & Lin et al.\cite{lin2023addressing} & Karel \\ \cline{2-6}
    & \multirow{3}{*}{\makecell{Explainable\\Policies}} & Policy Reasoning by FOL & PPO & Delfosse et al.\cite{delfosse2024interpretable} & Object-Centric Atari\\ \cline{3-6}
    & & \makecell{Policy Reasoning\\by Natural Language} & A2C & Peng et al.\cite{peng2022inherently} & Jericho \\ \cline{2-6}
    \hline
    \multirow{10}{*}{\makecell{Safety}}  & \multirow{5}{*}{\makecell{Logic\\Languages}}    & Safety by DFA Monitor & SARSA &  De Giacomo et al. \cite{de2019foundations} & \makecell{Breakout, Sapientino,\\Cocktail Party, MineCraft}\\ \cline{3-6}
    & & Safety by DFA Monitor & SARSA, Q-Learning, DQN & Alshiekh et al.\cite{alshiekh2018safe} & \makecell{GridWorld, Self-Driving Car,\\Water Tank, Pacman} \\ \cline{3-6}
    & & Safety by FOL & DQN & Zhang et al.\cite{zhang2019faster} & \makecell{Flappy Bird, Aircraft Shooting,\\Breakout, GridWorld} \\ \cline{3-6}
    & & Inferring Safety Monitor & DQN & Xu et al.\cite{xu38joint} & Letter, Maze \\ \cline{2-6}
    & \multirow{5}{*}{\makecell{Natural\\Languages}}  & Free-Form Constraint &  PPO & Prakash et al. \cite{prakash2020guiding} & MiniGrid\\ \cline{3-6}
    & & Free-Form Constraint & \makecell{Projection-Based\\Constrained Policy Optimization} & Yang et al. \cite{yang2021safe} & HazardWorld \\ \cline{3-6}
    & & Free-Form Constraint  & PPO & Lou et al. \cite{lou2024safe} & HazardWorld, SafetyGoal\\ \cline{3-6}
    & & Trajectory-Level Safety & PPO & Dong et al. \cite{dongtext} & HazardWorld, SafetyGoal \\ \cline{3-6}\
    & & Safety in MARL & PPO & Wang et al. \cite{wang2024safe} & LaMaSafe\\ \cline{3-6}
    \hline
    \hline
    \end{tabular}
    \label{tab:safety}
\end{table*}
\fi

\subsection{Interpretability}
Despite the impressive capabilities of DRL algorithms, their lack of interpretability is a significant concern. KRR techniques offer a more formal and interpretable way to enhance the interpretability of RL systems from various perspectives, ranging from rewards to policies. 

\begin{figure*}[!t]
    \centering
    \subfloat[]{
        \includegraphics[width=0.50\linewidth]{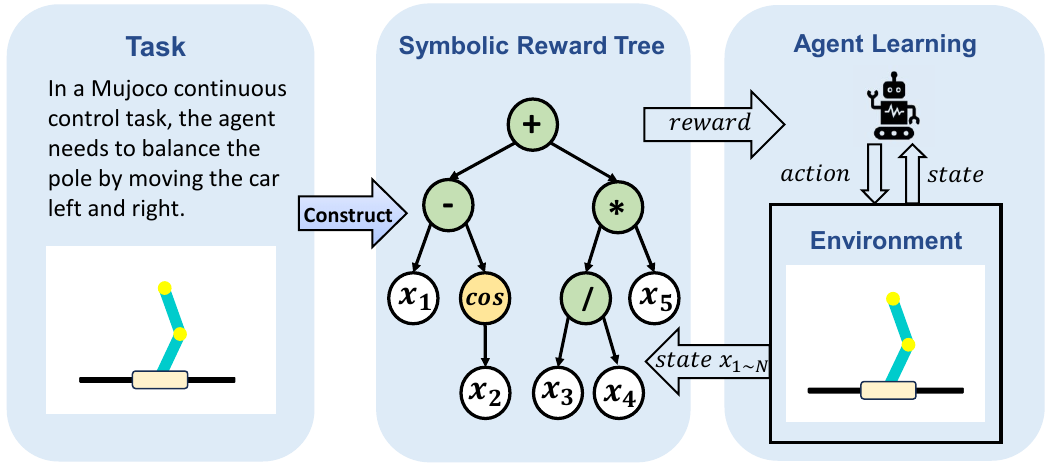}
        \label{fig:symbolic_reward}
    }
    \hfil
    \subfloat[]{
        \includegraphics[width=0.46\linewidth]{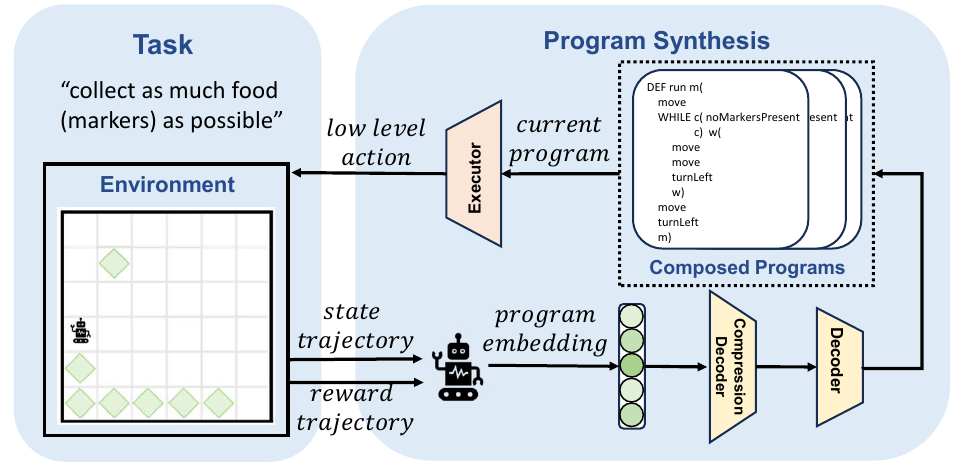}
        \label{fig:program_synthesis}
    }
    \caption{An illustration of
    (a) learning through symbolic reward functions; and
    (b) program synthesis.
    In (a), a symbolic reward tree is generated and modified for a Mujoco control task using evolutionary algorithms. In this tree, the leaf nodes represent the values for each dimension of the state vector $x_1,x_2,...,x_5$, while the non-leaf nodes act as arithmetic operators. The tree can be converted into a reward function through in-order traversal, resulting in the expression $reward=(x_1-cosx_2)+(x_3/x_4)*x_5$. 
    In (b), based on the state and reward trajectories of a task, the agent outputs a program embedding that is decoded into a program (programmatic policy). This program represents a high-level action, while low-level actions are executed by a predefined program executor. The agent continues to generate programs until the task is completed, and these generated programs are combined to form an overall policy for the task.
    }
    \label{fig:interpretability}
\end{figure*}

\textbf{Interpretable Rewards.}
Interpretable rewards can be generated by concise, human-readable reward functions expressed using a small set of symbolic arithmetic operators or programmatic languages. 
Works in this area can automatically obtain interpretable reward functions without the need to design them explicitly, as is done in aforementioned RM or TL approaches.
Sheikh et al.~\cite{sheikh2020learning} present a method that discovers dense rewards in the form of low-dimensional symbolic trees  to map the agent's observations to scalar rewards using basic arithmetic and logical operations, thus making them more tractable for analysis, as shown in Fig.~\ref{fig:symbolic_reward}. 
Bougie et al.~\cite{bougie2023interpretable} employ an inverse RL method to derive symbolic reward functions from expert data, and present a hierarchical representation of the reward function to accommodate scenarios with large state/action spaces.
Zhou et al.~\cite{zhou2022programmatic} propose a method that employs a generative adversarial learning framework to infer the optimal programmatic reward function from expert demonstrations.
While programmatic reward functions offer enhanced semantic expressiveness, identifying a suitable program can be quite difficult without a pre-existing program sketch. Obtaining such sketches often necessitates significant effort from experts.


\textbf{Symbolic Policies.}
Unlike the aforementioned methods that focus on making the rewards interpretable, symbolic policies aim to express RL policies in an interpretable manner using arithmetic operators and observation inputs.
Hein et al.~\cite{hein2018interpretable} propose a model-based RL method that learns a world model from trajectories and symbolic policies based on \textit{Genetic Programming} (GP). 
Bastani et al.~\cite{bastani2018verifiable} propose a method that extracts decision tree policies from pre-trained neural network policies, leveraging concepts from model compression and imitation learning.
Landajuela et al.~\cite{landajuela2021discovering} propose \textit{deep symbolic policy} to directly search the space of symbolic policies, as shown in Fig.~\ref{fig:symbolic_policy}. 
They use an autoregressive RNN to generate symbolic policies, and propose an ``anchoring'' algorithm to distill pre-trained neural network-based policies into fully symbolic policies in environments with multidimensional action spaces.

Although serving different purposes, the above studies all require pre-trained neural network policies to guide the generation of symbolic policies, resulting in additional computational effort.
Guo et al.~\cite{guo2024efficient} thus propose an efficient gradient-based framework that learns symbolic policies from scratch in an end-to-end manner, achieving performance that is comparable to or even surpasses that of previous works.
However, all these works still face the scalability issues of the symbolic policies in more challenging tasks with high-dimensional observations. When faced with complex tasks, these symbolic policies become extremely intricate, leading to a decrease in interpretability.

\textbf{Programmatic Policies.}
Recent research investigates the interpretability issues of RL through \textit{program synthesis}, representing policies in the form of programs.
In contrast to tree-structured symbolic policies, programmatic policies offer a broader semantic scope.
In specific, Verma et al.~\cite{verma2018programmatically} propose an approach to learn programmatic policies by distilling or imitating a pre-trained deep RL policy into short programs.
However, this approach usually relies on some form of supervised learning to mimic a pretrained policy, leading to catastrophic failure in certain cases. 
To address this limitation, Verma et al.~\cite{verma2019imitation} further frame the programmatic policy learning problem as a constrained optimization problem that employs constrained mirror descent techniques.  
Besides, there are some studies that have explored the use of ILP to enable RL agents to synthesize programmatic policies~\cite{jiang2019neural, cao2022galois}.
For example, Jiang et al.~\cite{jiang2019neural} propose an architecture that integrates \textit{Differentiable Inductive Logic Programming} (DILP) into RL.
This approach conducts logical deduction in a differentiable manner and extracts policies expressed in FOL.
Such FOL programs lack hierarchical logic. Then, Cao et al.~\cite{cao2022galois} introduce a new domain-specific language with hierarchical semantics and use a sketch-based program synthesis method based on DILP.

However, the generated programs in these works~\cite{verma2018programmatically, verma2019imitation} are limited to a set of predefined program templates or sketches, which restricts the scope of synthesizable programs and requires the design of specific templates. 
Additionally, \textit{program synthesis} using ILP~\cite{jiang2019neural, cao2022galois} is based on the assumption that a set of background facts is provided in advance.
To mitigate this limitation, Trivedi et al.~\cite{trivedi2021learning} propose a framework that learns a program embedding space and employ the cross-entropy method to identify the best program.
Building on this, Liu et al.~\cite{liu2023hierarchical} enhance the framework to enable out-of-distribution policy generation and improve reward assignment accuracy. 
As shown in Fig.~\ref{fig:program_synthesis}, their improved framework learns a meta-policy that composes a sequence of programs based on the learned embedding space, resembling the decision-making process of an RL agent. 
Drawing inspiration from state machine policies~\cite{inala2020synthesizing}, Lin et al.~\cite{lin2023addressing} further enhance this framework to tackle long-horizon tasks by incorporating state machines.
However, these works share the assumption that a program executor is pre-defined and does not take into account the interpretability of the program executor. Furthermore, these works limit their focus to deterministic environments, overlooking the uncertainties present in real-world scenarios.


While previous works have achieved interpretability in policies, they cannot clarify the importance of each input on its decision.
To achieve this objective, Delfosse et al.~\cite{delfosse2024interpretable} introduce a novel method that learns interpretable policies expressed in FOL and generates explanations for their decision-making.
Peng et al.~\cite{peng2022inherently} create an explainable agent which uses a knowledge graph-based state representation to provide step-by-step explanations for executed actions, as well as conducting post-hoc analyses to generate temporal explanations.





\subsection{Safety}
\begin{figure}[t]
    \centering
    \includegraphics[width=0.49\textwidth]{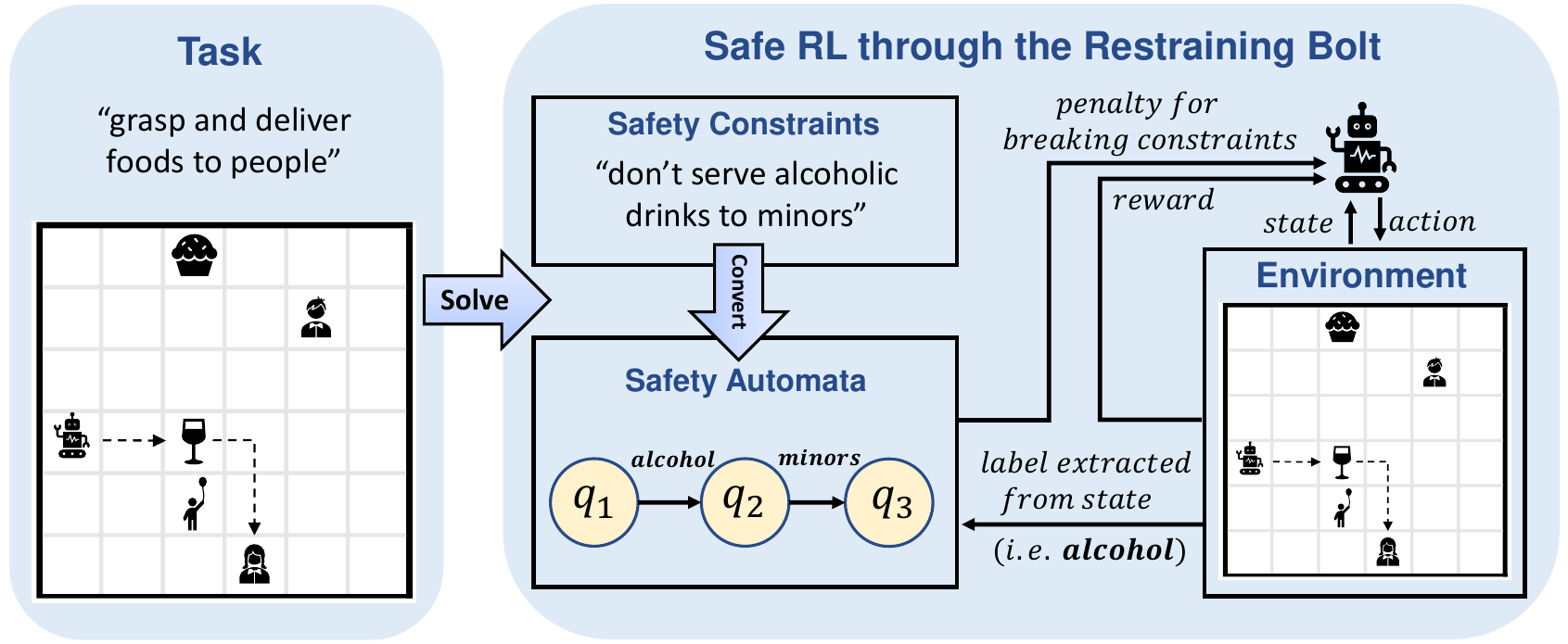}
    \caption{An illustration of imposing safety constraints by the \textit{restraining bolt}. In this scenario, the agent is tasked with picking up items and delivering them to people, earning a reward for each successful delivery. A key safety constraint provided by the user is: do not serve alcoholic drinks to minors. This constraint can be expressed as an LTL formula and converted into an equivalent automaton. As the agent interacts with the environment, it will earn rewards from the environment and receive additional rewards or penalties from the \textit{restraining bolt} through model checking.
    }
    \label{fig:restraining_bolt_example}
\end{figure}

\textbf{Logic Languages.}
Expressive logic formulas have also been used as constraints to guarantee safety during learning~\cite{alshiekh2018safe,jansen2018shielded,anderson2020neurosymbolic,junges2016safety,pathak2018verification,xu38joint,li2019formal,hunt2021verifiably,hasanbeig2020cautious,hasanbeig2020towards,hasanbeig2023certified}.
Most of this work employs TLs to describe the abstract states of agents, using mechanisms such as automata to determine whether the agents meet safety constraints.
In particular, De Giacomo et al.~\cite{de2019foundations} establish the foundations for the \textit{restraining bolt}.
In the proposed framework, when the agent's actions align with safety specifications, it receives additional rewards from the \textit{restraining bolt}, as illustrated in Fig.~\ref{fig:restraining_bolt_example}.
The concept of \textit{shield} is employed in~\cite{alshiekh2018safe} to synthesise a reactive system that ensures that the agent stays safe during and after learning. 
Actions that could violate the specified safety properties will be corrected by the \textit{shield}.
Building on the concept of \textit{shield}, subsequent works have expanded its application to various complex scenarios, including MARL~\cite{elsayed2021safe} and POMDP~\cite{carr2023safe}, to effectively address real-world tasks that are critical to safety~\cite{nikou2021symbolic,zhao2022safe}.
However, constructing both the \textit{restraining bolt} and the \textit{shield} requires specific TLs constraints, which are assumed to be known to the RL agent as specifications provided by the human user. In some cases, such specifications are implicit and difficult to define. 
To address this issue, Xu et al.~\cite{xu38joint} propose a joint learning framework for safety constraints and policies that relies on human feedback under unknown temporal logic constraints.  
Drawing inspiration from~\cite{tasse2020boolean}, the agent can make safe decisions by combining Q functions for rewards and constraints together.
However, this work still requires a considerable amount of effort to derive the manual feedback. 

While the aforementioned works utilize high-level knowledge to enhance RL agents in an indirect manner, such as constructing an automaton as a safety monitor, 
Zhang et al.~\cite{zhang2019faster} integrate rules directly into the learning process by proposing a framework of \textit{Rule-interposing Learning} (RIL) . With rules specified by first-order logic, RIL can not only accelerate the learning process, but also avoid catastrophic explorations, thus making the system relatively stable even during the very early stage of training. Moreover, given the rules are high-level and easy to interpret, they can be easily maintained, updated and shared with other similar tasks. 

\textbf{Natural Languages.}
It is more convenient and intuitive for users to convey constraints in natural language since high-level knowledge about safety constraints may be difficult to express in logical languages in some cases. 
Prakash et al.~\cite{prakash2020guiding} use a constraint checker to assign negative rewards to agents when they violate a constraint specified in structured language.
Instead of coupling rewards and constraints in~\cite{prakash2020guiding},
Yang et al.~\cite{yang2021safe} propose a modular architecture, in which a constraint interpreter converts language constraints into structured representations, while a policy network utilizes these representations to generate actions that minimize constraint violations.
However, the constraint interpreter needs to be trained from scratch, which demands additional computational resources.
Instead, Lou et al.~\cite{lou2024safe} propose a framework to use pre-trained LLMs to generate embeddings for language constraints and text-based observations, guiding the training of RL agents with costs derived from comparisons between these two kind of embeddings.
Dong et al.~\cite{dongtext} investigate a broader context known as trajectory-level textual constraints that can change over time within a trajectory, while Wang et al.~\cite{wang2024safe} utilize LLMs to adhere to safety constraints within an MARL setting.

\section{Challenges}
\label{sec:sixth}
Until now, we have surveyed the broad literature on applying various KRR techniques in addressing some key issues in RL. 
While various methods have been proposed and great progress has been made, there are still some everlasting challenges to be addressed for further research, which are briefly discussed below.

\subsection{Increasing the Spectrum of KRR/RL Methods}

The existing methods in the literature mainly employ formal declarative languages such as LTL for task specification or reasoning methods based on simple action-models for planning over actions. However, apart from these, there are still a variety of KRR concepts that are available for more comprehensive representation and reasoning of knowledge but have not been carefully combined with RL, such as \textit{production systems}, \textit{objects and frames}, \textit{inheritance networks}, and \textit{probabilistic causal models}~\cite{brachman2004knowledge}. How to model an RL problem by taking advantage of these powerful methods is still a relatively untouched issue in this area.

From the other perspective, the field of RL consists of various types of sub-topics such as MARL, MBRL and HRL, but the exiting KRR-driven RL methods have mainly focused on native model-free methods such as Q-learning and its function approximation versions. A continuing challenge is to investigate more types of RL methods using KRR methods for a better performance. An interesting direction can be MBRL, where a learned transition model embodying knowledge about the environment could be applied for generating more pseudo samples to improve learning efficiency, or reused by an agent in situations it has never experienced before. While there has been some work that jointly supports both logic-based and probabilistic representations of knowledge by associating probabilities with specific logic expressions (e.g., MLN), learning a probabilistic transition model with a logic structure under the RL framework is still a challenging issue that requires more research.
Another type of methods that have not been explicitly addressed include those under the partially observable settings. While this field itself has been well studied in the learning and planning community and has developed many exact and approximate algorithms, how to extend them to more powerful logic representations is still a challenging issue. A useful way to start is to make a connection with epistemic logic that is able to deal with modelling belief states. While there has been some progress, e.g., combining \textit{Situation Calculus} (SC) with epistemic aspects, more work still requires to be done along this line in the RL research.

\subsection{Analyzing Theoretical Properties}

Currently, most existing work still focuses on empirical evaluation on a specific learning task or domain. Development of theories on how and why some methods work is still in its initial phase. 
Theoretical advances are urgently needed to be able to understand better the benefits of using KRR in RL, as well as the interactions between various aspects, i.e., \textit{logic representation}, \textit{utility optimization} and \textit{uncertainty modelling}. Considering that the theoretical evaluation of sample complexity and convergence analysis is already a tricky problem in traditional RL research, analyzing its theoretical properties specified by logic can bring about extra challenges due to the atomic nature of logic expressions. Currently, work on this issue is limited to some initial convergence results and error bounds for specific systems. Moreover, given that using KRR in the learning process can generally be viewed as an abstraction over the original problem, it is feasible to derive useful theoretical properties for fixed abstraction levels. However, because the abstraction level can change during the learning process, it is largely an open question how to guarantee theoretical performance in these contexts, especially for high-dimensional environments
requiring function approximation.


The trade-off between more powerful representation and the increased cost of manipulating relational structures during learning should also be investigated. Using more expressive logic such as SC enables more powerful abstraction and thus lower space complexity, but at the same time is associated with an increased cost of manipulating and learning it. It is therefore necessary to provide formal theoretical analysis of comparing formal languages in terms of RL-specific criteria, such as \textit{sample complexity} and \textit{transfer efficiency}. Moreover, it is interesting to provide rigid theoretical evaluation on the trade-off between obtaining a stable abstraction using KRR and value convergence within an abstraction level.

\subsection{Integrating LLMs with KRR Techniques}
Most existing methods that employ KRR techniques require predefined logical formulas, rules, or knowledge bases, such as LTL task specifications or PDDL models. However, constructing accurate and comprehensive knowledge bases often requires significant expert experience and domain-specific knowledge and may not be comprehensive or accurate in complex domains. Currently, LLMs are showing great promise in enhancing RL agents, primarily by leveraging their vast, pre-trained knowledge base. This knowledge can be used to complement or even replace the need for explicitly defining task specifications or domain models in many cases.
Although there are some works~\cite{liu2022lang2ltl,mahdavi2024leveraging} that generate knowledge through LLM inference, systematically utilizing LLMs to enhance the RL agent’s ability to continuously update their knowledge and adapt to changing environments remains a significant unresolved challenge.

In addition, the sources of knowledge for real-world tasks are diverse and multimodal, including visual signals, textual descriptions, user feedback, and more. Current KRR methods have limited capabilities in handling heterogeneous data, particularly when it comes to unified logical reasoning after the integration of multimodal information. 
Multimodal LLMs offer a significant opportunity to address this challenge, ultimately bridging the gap between disparate data types and enhancing the overall robustness of RL systems.

\subsection{Applying in More Complex Domains}
A final important issue is regarding applications in more complex real-life domains. The existing research mainly works on simple discrete domains such as the Officeworld and Minecraft domain, or continuous domains such as the water world and robotic control, to show the effectiveness of the approaches. In order to fully show the benefits of KRRs for RL, it is necessary to apply the methods in solving larger problems for which naive representation methods are insufficient. However, scaling to more complex domains would be challenging due to the space of possible options to be considered, e.g., the size of the data to be reasoned with by the KRR methods, and the size of the state-action space to be considered by the RL methods. 

Another important issue is to scale the existing methods to problems involving multiple agents. 
Although there have already been some studies on this issue~\cite{cubuktepe2020policy,djeumou2020probabilistic}, most existing work simply focuses on multiagent task specifications using more expressive formal languages, but lack efficient reasoning mechanisms about strategic abilities among the agents. The foundational work on ATL~\cite{alur2002alternating} provides an efficient way of formalizing the concurrent interactions among agents~\cite{liu2020modal,xiong2016strategy}, and how to incorporate this mechanism into MARL and properly address the credit assignment and curse of dimensionality issues during learning is still an open problem that requires more evaluation.

\section{Conclusions}
Both RL and KRR aim at dealing with challenging AI problems in that RL focuses on interactive decision making in complex dynamic environments, while KRR provides a way of formally representing domain knowledge and reasoning over it. A paradigm that will prove fruitful for general AI realization is thus through building agents equipped with both reasoning capabilities powered by high-level knowledge representation and utilization, and explorative learning capabilities provided by RL. In recent years, the field of integrating KRR methods into RL has rapidly expanded, which can be witnessed by the increasing volume of work in this field. In this survey, we surveyed some of these studies that leverage the strengths of KRR to help addressing various problems in RL, including \emph{the efficiency}, \emph{the generalization}, as well as \emph{the safety and interpretability} issues, and briefly discussed several challenging open problems and possible directions for future work. Note that, due to page limit, the literature we reviewed in this survey is far from being complete, and we only selected several most notable studies for each category. We hope that this brief survey could inspire researchers from diverse fields in computer science, AI, statistics, logic, and cognitive science to continuously contribute to this exciting field.



\begin{thebibliography}{1}
\bibliographystyle{IEEEtran}

\bibitem{sutton2018reinforcement}
R.~S. Sutton and A.~G. Barto, \emph{Reinforcement learning: An introduction}.\hskip 1em plus 0.5em minus 0.4em\relax MIT press, 2018.

\bibitem{li2017deep}
Y.~Li, ``Deep reinforcement learning: An overview,'' \emph{arXiv preprint arXiv:1701.07274}, 2017.

\bibitem{yang2021exploration}
T.~Yang, H.~Tang, C.~Bai, and et~al., ``Exploration in deep reinforcement learning: a comprehensive survey,'' \emph{arXiv preprint arXiv:2109.06668}, 2021.

\bibitem{kirk2021survey}
R.~Kirk, A.~Zhang, E.~Grefenstette, and T.~Rockt{\"a}schel, ``A survey of generalisation in deep reinforcement learning,'' \emph{arXiv preprint arXiv:2111.09794}, 2021.

\bibitem{garcia2015comprehensive}
J.~Garc{\i}a and F.~Fern{\'a}ndez, ``A comprehensive survey on safe reinforcement learning,'' \emph{Journal of Machine Learning Research}, vol.~16, no.~1, pp. 1437--1480, 2015.

\bibitem{brachman2004knowledge}
R.~Brachman and H.~Levesque, \emph{Knowledge representation and reasoning}.\hskip 1em plus 0.5em minus 0.4em\relax Elsevier, 2004.

\bibitem{watkins1992q}
C.~J. Watkins and P.~Dayan, ``Q-learning,'' \emph{Machine learning}, vol.~8, no. 3-4, pp. 279--292, 1992.

\bibitem{arulkumaran2017deep}
K.~Arulkumaran, M.~P. Deisenroth, M.~Brundage, and A.~A. Bharath, ``Deep reinforcement learning: A brief survey,'' \emph{IEEE Signal Processing Magazine}, vol.~34, no.~6, pp. 26--38, 2017.

\bibitem{glanois2024survey}
C.~Glanois, P.~Weng, M.~Zimmer, D.~Li, T.~Yang, J.~Hao, and W.~Liu, ``A survey on interpretable reinforcement learning,'' \emph{Machine Learning}, pp. 1--44, 2024.

\bibitem{moerland2023model}
T.~M. Moerland, J.~Broekens, A.~Plaat, C.~M. Jonker \emph{et~al.}, ``Model-based reinforcement learning: A survey,'' \emph{Foundations and Trends{\textregistered} in Machine Learning}, vol.~16, no.~1, pp. 1--118, 2023.

\bibitem{pateria2021hierarchical}
S.~Pateria, B.~Subagdja, A.-h. Tan, and C.~Quek, ``Hierarchical reinforcement learning: A comprehensive survey,'' \emph{ACM Computing Surveys (CSUR)}, vol.~54, no.~5, pp. 1--35, 2021.

\bibitem{vithayathil2020survey}
N.~Vithayathil~Varghese and Q.~H. Mahmoud, ``A survey of multi-task deep reinforcement learning,'' \emph{Electronics}, vol.~9, no.~9, p. 1363, 2020.

\bibitem{khetarpal2022towards}
K.~Khetarpal, M.~Riemer, I.~Rish, and D.~Precup, ``Towards continual reinforcement learning: A review and perspectives,'' \emph{Journal of Artificial Intelligence Research}, vol.~75, pp. 1401--1476, 2022.

\bibitem{zhang2021multi}
K.~Zhang, Z.~Yang, and T.~Ba{\c{s}}ar, ``Multi-agent reinforcement learning: A selective overview of theories and algorithms,'' \emph{Handbook of Reinforcement Learning and Control}, pp. 321--384, 2021.

\bibitem{zambaldi2018relational}
V.~Zambaldi, D.~Raposo, A.~Santoro, V.~Bapst, Y.~Li, I.~Babuschkin, K.~Tuyls, D.~Reichert, T.~Lillicrap, E.~Lockhart \emph{et~al.}, ``Relational deep reinforcement learning,'' \emph{arXiv preprint arXiv:1806.01830}, 2018.

\bibitem{zhang2020survey}
S.~Zhang and M.~Sridharan, ``A survey of knowledge-based sequential decision making under uncertainty,'' \emph{arXiv preprint arXiv:2008.08548}, 2020.

\bibitem{pnueli1977temporal}
A.~Pnueli, ``The temporal logic of programs,'' in \emph{18th Annual Symposium on Foundations of Computer Science}, 1977, pp. 46--57.

\bibitem{mogavero2014reasoning}
F.~Mogavero, A.~Murano, G.~Perelli, and et~al., ``Reasoning about strategies: On the model-checking problem,'' \emph{ACM Transactions on Computational Logic}, vol.~15, no.~4, pp. 1--47, 2014.

\bibitem{lloyd2012foundations}
J.~W. Lloyd, \emph{Foundations of logic programming}.\hskip 1em plus 0.5em minus 0.4em\relax Springer Science \& Business Media, 2012.

\bibitem{de2015probabilistic}
L.~De~Raedt and A.~Kimmig, ``Probabilistic (logic) programming concepts,'' \emph{Machine Learning}, vol. 100, no.~1, pp. 5--47, 2015.

\bibitem{2006Markov}
M.~Richardson and P.~Domingos, ``Markov logic networks,'' \emph{Machine Learning}, vol.~62, no. 1-2, pp. 107--136, 2006.

\bibitem{1998PDDL}
M.~Ghallab, C.~Knoblock, D.~Wilkins, and et~al., ``Pddl - the planning domain definition language,'' 1998.

\bibitem{jothimurugan2019composable}
K.~Jothimurugan, R.~Alur, and O.~Bastani, ``A composable specification language for reinforcement learning tasks,'' in \emph{NeurIPS}, 2019, pp. 13\,041--13\,051.

\bibitem{icarte2018using}
R.~T. Icarte, T.~Klassen, R.~Valenzano, and S.~McIlraith, ``Using reward machines for high-level task specification and decomposition in reinforcement learning,'' in \emph{ICML}, 2018, pp. 2107--2116.

\bibitem{icarte2022reward}
R.~T. Icarte, T.~Q. Klassen, R.~Valenzano, and S.~A. McIlraith, ``Reward machines: Exploiting reward function structure in reinforcement learning,'' \emph{Journal of Artificial Intelligence Research}, vol.~73, pp. 173--208, 2022.

\bibitem{neary2020reward}
C.~Neary, Z.~Xu, B.~Wu, and U.~Topcu, ``Reward machines for cooperative multi-agent reinforcement learning,'' \emph{arXiv preprint arXiv:2007.01962}, 2020.

\bibitem{hu2021decentralized}
J.~Hu, Z.~Xu, W.~Wang, and et~al., ``Decentralized graph-based multi-agent reinforcement learning using reward machines,'' \emph{arXiv preprint arXiv:2110.00096}, 2021.

\bibitem{zheng2024multi}
X.~Zheng and C.~Yu, ``Multi-agent reinforcement learning with a hierarchy of reward machines,'' \emph{arXiv preprint arXiv:2403.07005}, 2024.

\bibitem{icarte2019learning}
R.~T. Icarte, E.~Waldie, T.~Klassen, and et~al., ``Learning reward machines for partially observable reinforcement learning,'' in \emph{NeurIPS}, 2019, pp. 15\,523--15\,534.

\bibitem{icarte2023learning}
R.~T. Icarte, T.~Q. Klassen, R.~Valenzano, M.~P. Castro, E.~Waldie, and S.~A. McIlraith, ``Learning reward machines: A study in partially observable reinforcement learning,'' \emph{Artificial Intelligence}, vol. 323, p. 103989, 2023.

\bibitem{xu2020joint}
Z.~Xu, I.~Gavran, and et~al., ``Joint inference of reward machines and policies for reinforcement learning,'' in \emph{ICAPS}, 2020, pp. 590--598.

\bibitem{furelos2020induction}
D.~Furelos-Blanco and et~al., ``Induction of subgoal automata for reinforcement learning,'' in \emph{AAAI}, 2020, pp. 3890--3897.

\bibitem{varricchione2023synthesising}
G.~Varricchione, N.~Alechina, M.~Dastani, and B.~Logan, ``Synthesising reward machines for cooperative multi-agent reinforcement learning,'' in \emph{European Conference on Multi-Agent Systems}.\hskip 1em plus 0.5em minus 0.4em\relax Springer, 2023, pp. 328--344.

\bibitem{ardon2023learning}
L.~Ardon, D.~Furelos-Blanco, and A.~Russo, ``Learning reward machines in cooperative multi-agent tasks,'' in \emph{International Conference on Autonomous Agents and Multiagent Systems}.\hskip 1em plus 0.5em minus 0.4em\relax Springer, 2023, pp. 43--59.

\bibitem{hatanaka2023reinforcement}
W.~Hatanaka, R.~Yamashina, and T.~Matsubara, ``Reinforcement learning of action and query policies with ltl instructions under uncertain event detector,'' \emph{IEEE Robotics and Automation Letters}, 2023.

\bibitem{li2024reward}
A.~C. Li, Z.~Chen, T.~Q. Klassen, P.~Vaezipoor, R.~T. Icarte, and S.~A. McIlraith, ``Reward machines for deep rl in noisy and uncertain environments,'' \emph{arXiv preprint arXiv:2406.00120}, 2024.

\bibitem{dohmen2022inferring}
T.~Dohmen, N.~Topper, G.~Atia, A.~Beckus, A.~Trivedi, and A.~Velasquez, ``Inferring probabilistic reward machines from non-markovian reward signals for reinforcement learning,'' in \emph{Proceedings of the International Conference on Automated Planning and Scheduling}, vol.~32, 2022, pp. 574--582.

\bibitem{corazza2022reinforcement}
J.~Corazza, I.~Gavran, and D.~Neider, ``Reinforcement learning with stochastic reward machines,'' in \emph{Proceedings of the AAAI Conference on Artificial Intelligence}, vol.~36, no.~6, 2022, pp. 6429--6436.

\bibitem{furelos2023hierarchies}
D.~Furelos-Blanco, M.~Law, A.~Jonsson, K.~Broda, and A.~Russo, ``Hierarchies of reward machines,'' in \emph{International Conference on Machine Learning}.\hskip 1em plus 0.5em minus 0.4em\relax PMLR, 2023, pp. 10\,494--10\,541.

\bibitem{levina2024numeric}
K.~Levina, N.~Pappas, A.~Karapantelakis, A.~V. Feljan, and J.~Seipp, ``Numeric reward machines,'' \emph{arXiv preprint arXiv:2404.19370}, 2024.

\bibitem{bourel2023exploration}
H.~Bourel, A.~Jonsson, O.-A. Maillard, and M.~S. Talebi, ``Exploration in reward machines with low regret,'' in \emph{International Conference on Artificial Intelligence and Statistics}.\hskip 1em plus 0.5em minus 0.4em\relax PMLR, 2023, pp. 4114--4146.

\bibitem{corazza2023expediting}
J.~Corazza, H.~P. Aria, D.~Neider, and Z.~Xu, ``Expediting reinforcement learning by incorporating temporal causal information,'' in \emph{Causal Representation Learning Workshop at NeurIPS 2023}.

\bibitem{sun2023less}
H.~Sun and F.~Wu, ``Less is more: Refining datasets for offline reinforcement learning with reward machines,'' in \emph{Proceedings of the 2023 International Conference on Autonomous Agents and Multiagent Systems (AAMAS)}, 2023, pp. 1239--1247.

\bibitem{koprulu2023reward}
C.~Koprulu and U.~Topcu, ``Reward-machine-guided, self-paced reinforcement learning,'' in \emph{Uncertainty in Artificial Intelligence}.\hskip 1em plus 0.5em minus 0.4em\relax PMLR, 2023, pp. 1121--1131.

\bibitem{li2017reinforcement}
X.~Li, C.~Vasile, and C.~Belta, ``Reinforcement learning with temporal logic rewards,'' in \emph{IROS}, 2017, pp. 3834--3839.

\bibitem{jothimurugan2021compositional}
K.~Jothimurugan, S.~Bansal, O.~Bastani, and R.~Alur, ``Compositional reinforcement learning from logical specifications,'' \emph{NeurIPS}, vol.~34, pp. 10\,026--10\,039, 2021.

\bibitem{camacho2019ltl}
A.~Camacho, R.~T. Icarte, T.~Q. Klassen, and et~al., ``Ltl and beyond: Formal languages for reward function specification in reinforcement learning.'' in \emph{IJCAI}, 2019, pp. 6065--6073.

\bibitem{yuan2019modular}
L.~Z. Yuan, M.~Hasanbeig, A.~Abate, and D.~Kroening, ``Modular deep reinforcement learning with temporal logic specifications,'' \emph{arXiv preprint arXiv:1909.11591}, 2019.

\bibitem{voloshin2023eventual}
C.~Voloshin, A.~Verma, and Y.~Yue, ``Eventual discounting temporal logic counterfactual experience replay,'' in \emph{International Conference on Machine Learning}.\hskip 1em plus 0.5em minus 0.4em\relax PMLR, 2023, pp. 35\,137--35\,150.

\bibitem{voloshin2022policy}
C.~Voloshin, H.~Le, S.~Chaudhuri, and Y.~Yue, ``Policy optimization with linear temporal logic constraints,'' \emph{Advances in Neural Information Processing Systems}, vol.~35, pp. 17\,690--17\,702, 2022.

\bibitem{cubuktepe2020policy}
M.~Cubuktepe, Z.~Xu, and U.~Topcu, ``Policy synthesis for factored mdps with graph temporal logic specifications,'' \emph{arXiv preprint arXiv:2001.09066}, 2020.

\bibitem{djeumou2020probabilistic}
F.~Djeumou, Z.~Xu, and U.~Topcu, ``Probabilistic swarm guidance subject to graph temporal logic specifications,'' in \emph{Robotics: Science and Systems (RSS)}, 2020.

\bibitem{muniraj2018enforcing}
D.~Muniraj, K.~G. Vamvoudakis, and M.~Farhood, ``Enforcing signal temporal logic specifications in multi-agent adversarial environments: A deep q-learning approach,'' in \emph{IEEE CDC}, 2018, pp. 4141--4146.

\bibitem{leon2020extended}
B.~G. Le{\'o}n and F.~Belardinelli, ``Extended markov games to learn multiple tasks in multi-agent reinforcement learning,'' \emph{arXiv preprint arXiv:2002.06000}, 2020.

\bibitem{den2022reinforcement}
F.~Den~Hengst, V.~Fran{\c{c}}ois-Lavet, M.~Hoogendoorn, and F.~van Harmelen, ``Reinforcement learning with option machines,'' in \emph{31st International Joint Conference on Artificial Intelligence, IJCAI 2022}.\hskip 1em plus 0.5em minus 0.4em\relax International Joint Conferences on Artificial Intelligence Organization, 2022, pp. 2909--2915.

\bibitem{leonetti2016synthesis}
M.~Leonetti, L.~Iocchi, and P.~Stone, ``A synthesis of automated planning and reinforcement learning for efficient, robust decision-making,'' \emph{Artificial Intelligence}, vol. 241, pp. 103--130, 2016.

\bibitem{ferreira2018method}
L.~A. Ferreira, R.~A. Bianchi, P.~E. Santos, and R.~L. de~Mantaras, ``A method for the online construction of the set of states of a markov decision process using answer set programming,'' in \emph{International Conference on Industrial, Engineering and Other Applications of Applied Intelligent Systems}.\hskip 1em plus 0.5em minus 0.4em\relax Springer, 2018, pp. 3--15.

\bibitem{mitchener2022detect}
L.~Mitchener, D.~Tuckey, M.~Crosby, and A.~Russo, ``Detect, understand, act: A neuro-symbolic hierarchical reinforcement learning framework,'' \emph{Machine Learning}, vol. 111, no.~4, pp. 1523--1549, 2022.

\bibitem{nickles2012integrating}
M.~Nickles, ``Integrating relational reinforcement learning with reasoning about actions and change,'' in \emph{Inductive Logic Programming: 21st International Conference, ILP 2011, Windsor Great Park, UK, July 31--August 3, 2011, Revised Selected Papers 21}.\hskip 1em plus 0.5em minus 0.4em\relax Springer, 2012, pp. 255--269.

\bibitem{yang2018peorl}
F.~Yang, D.~Lyu, B.~Liu, and S.~Gustafson, ``Peorl: Integrating symbolic planning and hierarchical reinforcement learning for robust decision-making,'' \emph{arXiv preprint arXiv:1804.07779}, 2018.

\bibitem{lyu2019sdrl}
D.~Lyu, F.~Yang, B.~Liu, and S.~Gustafson, ``Sdrl: interpretable and data-efficient deep reinforcement learning leveraging symbolic planning,'' in \emph{AAAI}, 2019, pp. 2970--2977.

\bibitem{illanes2020symbolic}
L.~Illanes, X.~Yan, R.~T. Icarte, and S.~A. McIlraith, ``Symbolic plans as high-level instructions for reinforcement learning,'' in \emph{ICAPS}, 2020, pp. 540--550.

\bibitem{kokel2021reprel}
H.~Kokel, A.~Manoharan, S.~Natarajan, B.~Ravindran, and P.~Tadepalli, ``Reprel: Integrating relational planning and reinforcement learning for effective abstraction,'' in \emph{Proceedings of the International Conference on Automated Planning and Scheduling}, vol.~31, 2021, pp. 533--541.

\bibitem{kokel2022hybrid}
H.~Kokel, N.~Prabhakar, B.~Ravindran, E.~Blasch, P.~Tadepalli, and S.~Natarajan, ``Hybrid deep reprel: Integrating relational planning and reinforcement learning for information fusion,'' in \emph{2022 25th International Conference on Information Fusion (FUSION)}.\hskip 1em plus 0.5em minus 0.4em\relax IEEE, 2022, pp. 1--8.

\bibitem{toro2018teaching}
R.~Toro~Icarte, T.~Q. Klassen, and et~al., ``Teaching multiple tasks to an rl agent using ltl,'' in \emph{AAMAS}, 2018, pp. 452--461.

\bibitem{brafman2018ltlf}
R.~Brafman, G.~De~Giacomo, and F.~Patrizi, ``Ltlf/ldlf non-markovian rewards,'' in \emph{AAAI}, vol.~32, no.~1, 2018.

\bibitem{bozkurt2020control}
A.~K. Bozkurt, Y.~Wang, and et~al., ``Control synthesis from linear temporal logic specifications using model-free reinforcement learning,'' in \emph{IEEE ICRA}, 2020, pp. 10\,349--10\,355.

\bibitem{de2019foundations}
G.~De~Giacomo, L.~Iocchi, and et~al., ``Foundations for restraining bolts: Reinforcement learning with ltlf/ldlf restraining specifications,'' in \emph{ICAPS}, 2019, pp. 128--136.

\bibitem{aksaray2016q}
D.~Aksaray, A.~Jones, Z.~Kong, M.~Schwager, and C.~Belta, ``Q-learning for robust satisfaction of signal temporal logic specifications,'' in \emph{IEEE 55th CDC}, 2016, pp. 6565--6570.

\bibitem{hasanbeig2018logically}
M.~Hasanbeig, A.~Abate, and D.~Kroening, ``Logically-constrained reinforcement learning,'' \emph{arXiv preprint arXiv:1801.08099}, 2018.

\bibitem{hasanbeig2019reinforcement}
M.~Hasanbeig and et~al., ``Reinforcement learning for temporal logic control synthesis with probabilistic satisfaction guarantees,'' in \emph{CDC}, 2019, pp. 5338--5343.

\bibitem{littman2017environment}
M.~L. Littman, U.~Topcu, J.~Fu, C.~Isbell, M.~Wen, and J.~MacGlashan, ``Environment-independent task specifications via gltl,'' \emph{arXiv preprint arXiv:1704.04341}, 2017.

\bibitem{jiang2021temporal}
Y.~Jiang, S.~Bharadwaj, B.~Wu, and et~al., ``Temporal-logic-based reward shaping for continuing reinforcement learning tasks,'' in \emph{AAAI}, 2021, pp. 7995--8003.

\bibitem{sickert2016limit}
S.~Sickert, J.~Esparza, S.~Jaax, and J.~K{\v{r}}et{\'\i}nsk{\`y}, ``Limit-deterministic b{\"u}chi automata for linear temporal logic,'' in \emph{International Conference on Computer Aided Verification}.\hskip 1em plus 0.5em minus 0.4em\relax Springer, 2016, pp. 312--332.

\bibitem{le2024reinforcement}
X.-B. Le, D.~Wagner, L.~Witzman, A.~Rabinovich, and L.~Ong, ``Reinforcement learning with ltl and $\omega$-regular objectives via optimality-preserving translation to average rewards,'' \emph{arXiv preprint arXiv:2410.12175}, 2024.

\bibitem{hammond2021multi}
L.~Hammond, A.~Abate, J.~Gutierrez, and M.~Wooldridge, ``Multi-agent reinforcement learning with temporal logic specifications,'' in \emph{AAMAS}, 2021, pp. 583--592.

\bibitem{zhu2024decomposing}
C.~Zhu, W.~Si, J.~Zhu, and Z.~Jiang, ``Decomposing temporal equilibrium strategy for coordinated distributed multi-agent reinforcement learning,'' in \emph{Proceedings of the AAAI Conference on Artificial Intelligence}, vol.~38, no.~16, 2024, pp. 17\,618--17\,627.

\bibitem{terashima2024reward}
K.~Terashima, K.~Kobayashi, and Y.~Yamashita, ``On reward distribution in reinforcement learning of multi-agent surveillance systems with temporal logic specifications,'' \emph{Advanced Robotics}, vol.~38, no.~6, pp. 386--397, 2024.

\bibitem{ShahKSL18bayesian}
A.~Shah, P.~Kamath, and et~al., ``Bayesian inference of temporal task specifications from demonstrations,'' in \emph{NeurIPS}, 2018, pp. 3808--3817.

\bibitem{KimMSAS19bayesian}
J.~Kim, C.~Muise, A.~Shah, and et~al., ``Bayesian inference of linear temporal logic specifications for contrastive explanations,'' in \emph{IJCAI}, 2019, pp. 5591--5598.

\bibitem{abate2023learning}
A.~Abate, Y.~Almulla, J.~Fox, D.~Hyland, and M.~Wooldridge, ``Learning task automata for reinforcement learning using hidden markov models,'' in \emph{ECAI 2023}.\hskip 1em plus 0.5em minus 0.4em\relax IOS Press, 2023, pp. 3--10.

\bibitem{CamachoM19learning}
A.~Camacho and S.~A. McIlraith, ``Learning interpretable models expressed in linear temporal logic,'' in \emph{ICAPS}, 2019, pp. 621--630.

\bibitem{LuoLDWPZ22Bridging}
W.~Luo, P.~Liang, J.~Du, H.~Wan, B.~Peng, and D.~Zhang, ``Bridging ltlf inference to {GNN} inference for learning ltlf formulae,'' in \emph{AAAI}, 2022, pp. 9849--9857.

\bibitem{guan2022leveraging}
L.~Guan, S.~Sreedharan, and S.~Kambhampati, ``Leveraging approximate symbolic models for reinforcement learning via skill diversity,'' in \emph{International Conference on Machine Learning}.\hskip 1em plus 0.5em minus 0.4em\relax PMLR, 2022, pp. 7949--7967.

\bibitem{DBLP:journals/nature/SchrittwieserAH20}
J.~Schrittwieser, I.~Antonoglou, T.~Hubert, and et~al., ``Mastering atari, go, chess and shogi by planning with a learned model,'' \emph{Nat.}, vol. 588, no. 7839, pp. 604--609, 2020. [Online]. Available: \url{https://doi.org/10.1038/s41586-020-03051-4}

\bibitem{DBLP:conf/nips/Yin22}
Z.-H. Yin, W.~Ye, Q.~Chen, and Y.~Gao, ``Planning for sample efficient imitation learning,'' in \emph{NeurIPS}, 2022.

\bibitem{jin2022creativity}
M.~Jin, Z.~Ma, K.~Jin, H.~H. Zhuo, C.~Chen, and C.~Yu, ``Creativity of ai: Automatic symbolic option discovery for facilitating deep reinforcement learning,'' in \emph{AAAI}, vol.~36, no.~6, 2022, pp. 7042--7050.

\bibitem{wu2022models}
Z.~Wu, C.~Yu, and et~al., ``Models as agents: Optimizing multi-step predictions of interactive local models in model-based multi-agent reinforcement learning,'' in \emph{AAAI}, 2023.

\bibitem{wu2022Plan}
Z.~Wu, C.~Yu, C.~Chen, J.~Hao, and H.~H. Zhuo, ``Plan to predict: Learning an uncertainty-foreseeing model for model-based reinforcement learning,'' in \emph{NeurIPS}, 2022.

\bibitem{mahdavi2024leveraging}
S.~Mahdavi, R.~Aoki, K.~Tang, and Y.~Cao, ``Leveraging environment interaction for automated pddl translation and planning with large language models,'' \emph{arXiv preprint arXiv:2407.12979}, 2024.

\bibitem{guan2023leveraging}
L.~Guan, K.~Valmeekam, S.~Sreedharan, and S.~Kambhampati, ``Leveraging pre-trained large language models to construct and utilize world models for model-based task planning,'' \emph{Advances in Neural Information Processing Systems}, vol.~36, pp. 79\,081--79\,094, 2023.

\bibitem{smirnov2024generating}
P.~Smirnov, F.~Joublin, A.~Ceravola, and M.~Gienger, ``Generating consistent pddl domains with large language models,'' \emph{arXiv preprint arXiv:2404.07751}, 2024.

\bibitem{landajuela2021discovering}
M.~Landajuela, B.~K. Petersen, S.~Kim, and et~al., ``Discovering symbolic policies with deep reinforcement learning,'' in \emph{ICML}, 2021, pp. 5979--5989.

\bibitem{khetarpal2020towards}
K.~Khetarpal, M.~Riemer, I.~Rish, and D.~Precup, ``Towards continual reinforcement learning: A review and perspectives,'' \emph{arXiv preprint arXiv:2012.13490}, 2020.

\bibitem{mihalkova2007mapping}
L.~Mihalkova, T.~Huynh, and R.~J. Mooney, ``Mapping and revising markov logic networks for transfer learning,'' in \emph{AAAI}, vol.~7, 2007, pp. 608--614.

\bibitem{mihalkova2008transfer}
L.~Mihalkova and R.~J. Mooney, ``Transfer learning by mapping with minimal target data,'' in \emph{Proceedings of the AAAI-08 workshop on transfer learning for complex tasks}, 2008, pp. 31--36.

\bibitem{torrey2010policy}
L.~Torrey and J.~Shavlik, ``Policy transfer via markov logic networks,'' in \emph{ILP}, 2010, pp. 234--248.

\bibitem{mansour2017recent}
R.~F. Mansour and S.~Hosni, ``Recent advances in markov logic networks,'' \emph{Indian Journal of Science and Technology}, vol.~10, p.~19, 2017.

\bibitem{li2017automata}
X.~Li, Y.~Ma, and C.~Belta, ``Automata-guided hierarchical reinforcement learning for skill composition,'' \emph{arXiv preprint arXiv:1711.00129}, 2017.

\bibitem{leon2020systematic}
B.~G. Leon, M.~Shanahan, and F.~Belardinelli, ``Systematic generalisation through task temporal logic and deep reinforcement learning,'' \emph{arXiv preprint arXiv:2006.08767}, 2020.

\bibitem{xu2024generalization2}
D.~Xu and F.~Fekri, ``Generalization of compositional tasks with logical specification via implicit planning,'' \emph{arXiv preprint arXiv:2410.09686}, 2024.

\bibitem{liu2024skill}
J.~X. Liu, A.~Shah, E.~Rosen, M.~Jia, G.~Konidaris, and S.~Tellex, ``Skill transfer for temporal task specification,'' in \emph{2024 IEEE International Conference on Robotics and Automation (ICRA)}.\hskip 1em plus 0.5em minus 0.4em\relax IEEE, 2024, pp. 2535--2541.

\bibitem{azran2024contextual}
G.~Azran, M.~H. Danesh, S.~V. Albrecht, and S.~Keren, ``Contextual pre-planning on reward machine abstractions for enhanced transfer in deep reinforcement learning,'' in \emph{Proceedings of the AAAI Conference on Artificial Intelligence}, vol.~38, no.~10, 2024, pp. 10\,953--10\,961.

\bibitem{xu2019transfer}
Z.~Xu and U.~Topcu, ``Transfer of temporal logic formulas in reinforcement learning,'' in \emph{IJCAI}, 2019, pp. 4010--4018.

\bibitem{kuo2020encoding}
Y.-L. Kuo, B.~Katz, and A.~Barbu, ``Encoding formulas as deep networks: Reinforcement learning for zero-shot execution of ltl formulas,'' in \emph{IEEE/RSJ IROS}, 2020, pp. 5604--5610.

\bibitem{vaezipoor2021ltl2action}
P.~Vaezipoor, A.~C. Li, R.~A.~T. Icarte, and S.~A. Mcilraith, ``Ltl2action: Generalizing ltl instructions for multi-task rl,'' in \emph{ICML}, 2021, pp. 10\,497--10\,508.

\bibitem{leon2021nutshell}
B.~G. Le{\'o}n, M.~Shanahan, and F.~Belardinelli, ``In a nutshell, the human asked for this: Latent goals for following temporal specifications,'' \emph{arXiv preprint arXiv:2110.09461}, 2021.

\bibitem{van2019composing}
B.~Van~Niekerk, S.~James, A.~Earle, and B.~Rosman, ``Composing value functions in reinforcement learning,'' in \emph{ICML}, 2019, pp. 6401--6409.

\bibitem{tasse2020boolean}
G.~N. Tasse, S.~James, and B.~Rosman, ``A boolean task algebra for reinforcement learning,'' \emph{arXiv preprint arXiv:2001.01394}, 2020.

\bibitem{tasse2020logical}
------, ``Logical composition in lifelong reinforcement learning,'' in \emph{4th Lifelong Machine Learning Workshop at ICML 2020}, 2020.

\bibitem{tasse2021generalisation}
------, ``Generalisation in lifelong reinforcement learning through logical composition,'' in \emph{ICML}, 2021.

\bibitem{tasseskill}
G.~N. Tasse, D.~Jarvis, S.~James, and B.~Rosman, ``Skill machines: Temporal logic skill composition in reinforcement learning,'' in \emph{The Twelfth International Conference on Learning Representations}.

\bibitem{qiu2024instructing}
W.~Qiu, W.~Mao, and H.~Zhu, ``Instructing goal-conditioned reinforcement learning agents with temporal logic objectives,'' \emph{Advances in Neural Information Processing Systems}, vol.~36, 2024.

\bibitem{xu2024generalization}
D.~Xu and F.~Fekri, ``Generalization of temporal logic tasks via future dependent options,'' \emph{Machine Learning}, pp. 1--32, 2024.

\bibitem{zheng2022lifelong}
X.~Zheng, C.~Yu, and M.~Zhang, ``Lifelong reinforcement learning with temporal logic formulas and reward machines,'' \emph{Knowledge-Based Systems}, vol. 257, p. 109650, 2022.

\bibitem{kuric2024planning}
D.~Kuric, G.~Infante, V.~G{\'o}mez, A.~Jonsson, and H.~van Hoof, ``Planning with a learned policy basis to optimally solve complex tasks,'' in \emph{Proceedings of the International Conference on Automated Planning and Scheduling}, vol.~34, 2024, pp. 333--341.

\bibitem{hanjie2021grounding}
A.~W. Hanjie, V.~Y. Zhong, and K.~Narasimhan, ``Grounding language to entities and dynamics for generalization in reinforcement learning,'' in \emph{International Conference on Machine Learning}.\hskip 1em plus 0.5em minus 0.4em\relax PMLR, 2021, pp. 4051--4062.

\bibitem{cao2024exploring}
L.~Cao, C.~Wang, J.~Qi, and Y.~Peng, ``Exploring into the unseen: Enhancing language-conditioned policy generalization with behavioral information,'' \emph{Cyborg and Bionic Systems}, vol.~5, p. 0084, 2024.

\bibitem{bing2023meta}
Z.~Bing, A.~Koch, X.~Yao, K.~Huang, and A.~Knoll, ``Meta-reinforcement learning via language instructions,'' in \emph{2023 IEEE International Conference on Robotics and Automation (ICRA)}.\hskip 1em plus 0.5em minus 0.4em\relax IEEE, 2023, pp. 5985--5991.

\bibitem{quartey2023exploiting}
B.~Quartey, A.~Shah, and G.~Konidaris, ``Exploiting contextual structure to generate useful auxiliary tasks,'' \emph{arXiv preprint arXiv:2303.05038}, 2023.

\bibitem{mumuni2022data}
A.~Mumuni and F.~Mumuni, ``Data augmentation: A comprehensive survey of modern approaches,'' \emph{Array}, vol.~16, p. 100258, 2022.

\bibitem{yalcinkaya2024compositional}
B.~Yalcinkaya, N.~Lauffer, M.~Vazquez-Chanlatte, and S.~A. Seshia, ``Compositional automata embeddings for goal-conditioned reinforcement learning,'' \emph{arXiv preprint arXiv:2411.00205}, 2024.

\bibitem{cao2024survey}
Y.~Cao, H.~Zhao, Y.~Cheng, T.~Shu, Y.~Chen, G.~Liu, G.~Liang, J.~Zhao, J.~Yan, and Y.~Li, ``Survey on large language model-enhanced reinforcement learning: Concept, taxonomy, and methods,'' \emph{arXiv preprint arXiv:2404.00282}, 2024.

\bibitem{rashidi2024survey}
A.~Rashidi~Laleh and M.~Nili~Ahmadabadi, ``A survey on enhancing reinforcement learning in complex environments: Insights from human and llm feedback,'' \emph{arXiv e-prints}, pp. arXiv--2411, 2024.

\bibitem{rocamonde2023vision}
J.~Rocamonde, V.~Montesinos, E.~Nava, E.~Perez, and D.~Lindner, ``Vision-language models are zero-shot reward models for reinforcement learning,'' \emph{arXiv preprint arXiv:2310.12921}, 2023.

\bibitem{yu2023b}
Z.~Yu, Y.~Tao, L.~Chen, T.~Sun, and H.~Yang, ``B-coder: Value-based deep reinforcement learning for program synthesis.'' \emph{CoRR}, 2023.

\bibitem{zhao2023test}
S.~Zhao, X.~Wang, L.~Zhu, and Y.~Yang, ``Test-time adaptation with clip reward for zero-shot generalization in vision-language models,'' \emph{arXiv preprint arXiv:2305.18010}, 2023.

\bibitem{kiran2021deep}
B.~R. Kiran, I.~Sobh, V.~Talpaert, P.~Mannion, A.~A. Al~Sallab, S.~Yogamani, and P.~P{\'e}rez, ``Deep reinforcement learning for autonomous driving: A survey,'' \emph{IEEE Transactions on Intelligent Transportation Systems}, 2021.

\bibitem{yu2021reinforcement}
C.~Yu, J.~Liu, S.~Nemati, and G.~Yin, ``Reinforcement learning in healthcare: A survey,'' \emph{ACM Computing Surveys (CSUR)}, vol.~55, no.~1, pp. 1--36, 2021.

\bibitem{nian2020review}
R.~Nian, J.~Liu, and B.~Huang, ``A review on reinforcement learning: Introduction and applications in industrial process control,'' \emph{Computers \& Chemical Engineering}, vol. 139, p. 106886, 2020.

\bibitem{sheikh2020learning}
H.~Sheikh, S.~Khadka, S.~Miret, and S.~Majumdar, ``Learning intrinsic symbolic rewards in reinforcement learning,'' \emph{arXiv preprint arXiv:2010.03694}, 2020.

\bibitem{bougie2023interpretable}
N.~Bougie, T.~Onishi, and Y.~Tsuruoka, ``Interpretable imitation learning with symbolic rewards,'' \emph{ACM Transactions on Intelligent Systems and Technology}, vol.~15, no.~1, pp. 1--34, 2023.

\bibitem{zhou2022programmatic}
W.~Zhou and W.~Li, ``Programmatic reward design by example,'' in \emph{Proceedings of the AAAI Conference on Artificial Intelligence}, vol.~36, no.~8, 2022, pp. 9233--9241.

\bibitem{hein2018interpretable}
D.~Hein, S.~Udluft, and et~al, ``Interpretable policies for reinforcement learning by genetic programming,'' \emph{Engineering Applications of Artificial Intelligence}, vol.~76, pp. 158--169, 2018.

\bibitem{bastani2018verifiable}
O.~Bastani, Y.~Pu, and A.~Solar-Lezama, ``Verifiable reinforcement learning via policy extraction,'' \emph{Advances in neural information processing systems}, vol.~31, 2018.

\bibitem{guo2024efficient}
J.~Guo, R.~Zhang, S.~Peng, Q.~Yi, X.~Hu, R.~Chen, Z.~Du, L.~Li, Q.~Guo, Y.~Chen \emph{et~al.}, ``Efficient symbolic policy learning with differentiable symbolic expression,'' \emph{Advances in Neural Information Processing Systems}, vol.~36, 2024.

\bibitem{verma2018programmatically}
A.~Verma, V.~Murali, R.~Singh, P.~Kohli, and S.~Chaudhuri, ``Programmatically interpretable reinforcement learning,'' in \emph{ICML}, 2018, pp. 5045--5054.

\bibitem{verma2019imitation}
A.~Verma, H.~Le, and et~al., ``Imitation-projected programmatic reinforcement learning,'' \emph{NeurIPS}, vol.~32, 2019.

\bibitem{jiang2019neural}
Z.~Jiang and S.~Luo, ``Neural logic reinforcement learning,'' in \emph{International conference on machine learning}.\hskip 1em plus 0.5em minus 0.4em\relax PMLR, 2019, pp. 3110--3119.

\bibitem{cao2022galois}
Y.~Cao, Z.~Li, T.~Yang, H.~Zhang, Y.~Zheng, Y.~Li, J.~Hao, and Y.~Liu, ``Galois: boosting deep reinforcement learning via generalizable logic synthesis,'' \emph{Advances in Neural Information Processing Systems}, vol.~35, pp. 19\,930--19\,943, 2022.

\bibitem{trivedi2021learning}
D.~Trivedi, J.~Zhang, S.-H. Sun, and J.~J. Lim, ``Learning to synthesize programs as interpretable and generalizable policies,'' \emph{Advances in neural information processing systems}, vol.~34, pp. 25\,146--25\,163, 2021.

\bibitem{liu2023hierarchical}
G.-T. Liu, E.-P. Hu, P.-J. Cheng, H.-Y. Lee, and S.-H. Sun, ``Hierarchical programmatic reinforcement learning via learning to compose programs,'' in \emph{International Conference on Machine Learning}.\hskip 1em plus 0.5em minus 0.4em\relax PMLR, 2023, pp. 21\,672--21\,697.

\bibitem{lin2023addressing}
Y.-A. Lin, C.-T. Lee, G.-T. Liu, P.-J. Cheng, and S.-H. Sun, ``Addressing long-horizon tasks by integrating program synthesis and state machines,'' \emph{arXiv preprint arXiv:2311.15960}, 2023.

\bibitem{delfosse2024interpretable}
Q.~Delfosse, H.~Shindo, D.~Dhami, and K.~Kersting, ``Interpretable and explainable logical policies via neurally guided symbolic abstraction,'' \emph{Advances in Neural Information Processing Systems}, vol.~36, 2024.

\bibitem{peng2022inherently}
X.~Peng, M.~Riedl, and P.~Ammanabrolu, ``Inherently explainable reinforcement learning in natural language,'' \emph{Advances in Neural Information Processing Systems}, vol.~35, pp. 16\,178--16\,190, 2022.

\bibitem{alshiekh2018safe}
M.~Alshiekh, R.~Bloem, R.~Ehlers, and et~al., ``Safe reinforcement learning via shielding,'' in \emph{AAAI}, 2018.

\bibitem{zhang2019faster}
H.~Zhang, Z.~Gao, Y.~Zhou, and et~al., ``Faster and safer training by embedding high-level knowledge into deep reinforcement learning,'' \emph{arXiv preprint arXiv:1910.09986}, 2019.

\bibitem{xu38joint}
D.~Xu and F.~Fekri, ``Joint learning of temporal logic constraints and policy in rl with human feedbacks,'' \emph{methods}, vol.~38, no.~31, p.~9.

\bibitem{prakash2020guiding}
B.~Prakash, N.~R. Waytowich, A.~Ganesan, T.~Oates, and T.~Mohsenin, ``Guiding safe reinforcement learning policies using structured language constraints.'' in \emph{SafeAI@ AAAI}, 2020, pp. 153--161.

\bibitem{yang2021safe}
T.-Y. Yang, M.~Y. Hu, Y.~Chow, P.~J. Ramadge, and K.~Narasimhan, ``Safe reinforcement learning with natural language constraints,'' \emph{Advances in Neural Information Processing Systems}, vol.~34, pp. 13\,794--13\,808, 2021.

\bibitem{lou2024safe}
X.~Lou, J.~Zhang, Z.~Wang, K.~Huang, and Y.~Du, ``Safe reinforcement learning with free-form natural language constraints and pre-trained language models,'' \emph{arXiv preprint arXiv:2401.07553}, 2024.

\bibitem{dongtext}
P.~Dong, T.~Zhu, Y.~Qiu, H.~Zhou, and J.~Li, ``From text to trajectory: Exploring complex constraint representation and decomposition in safe reinforcement learning,'' in \emph{The Thirty-eighth Annual Conference on Neural Information Processing Systems}.

\bibitem{wang2024safe}
Z.~Wang, M.~Fang, T.~Tomilin, F.~Fang, and Y.~Du, ``Safe multi-agent reinforcement learning with natural language constraints,'' \emph{arXiv preprint arXiv:2405.20018}, 2024.

\bibitem{rocha2024program}
F.~M. Rocha, I.~Dutra, and V.~S. Costa, ``Program synthesis using inductive logic programming for the abstraction and reasoning corpus,'' \emph{arXiv preprint arXiv:2405.06399}, 2024.

\bibitem{inala2020synthesizing}
J.~P. Inala, O.~Bastani, Z.~Tavares, and A.~Solar-Lezama, ``Synthesizing programmatic policies that inductively generalize,'' in \emph{ICLR}, 2020.

\bibitem{jansen2018shielded}
N.~Jansen, B.~K{\"o}nighofer, S.~Junges, and R.~Bloem, ``Shielded decision-making in mdps,'' \emph{arXiv preprint arXiv:1807.06096}, 2018.

\bibitem{anderson2020neurosymbolic}
G.~Anderson, A.~Verma, I.~Dillig, and S.~Chaudhuri, ``Neurosymbolic reinforcement learning with formally verified exploration,'' \emph{NeurIPS}, vol.~33, pp. 6172--6183, 2020.

\bibitem{junges2016safety}
S.~Junges, N.~Jansen, C.~Dehnert, U.~Topcu, and J.-P. Katoen, ``Safety-constrained reinforcement learning for mdps,'' in \emph{ICTA}, 2016, pp. 130--146.

\bibitem{pathak2018verification}
S.~Pathak, L.~Pulina, and A.~Tacchella, ``Verification and repair of control policies for safe reinforcement learning,'' \emph{Applied Intelligence}, vol.~48, no.~4, pp. 886--908, 2018.

\bibitem{li2019formal}
X.~Li, Z.~Serlin, G.~Yang, and C.~Belta, ``A formal methods approach to interpretable reinforcement learning for robotic planning,'' \emph{Science Robotics}, vol.~4, no.~37, 2019.

\bibitem{hunt2021verifiably}
N.~Hunt, N.~Fulton, S.~Magliacane, T.~N. Hoang, S.~Das, and A.~Solar-Lezama, ``Verifiably safe exploration for end-to-end reinforcement learning,'' in \emph{Proceedings of the 24th International Conference on Hybrid Systems: Computation and Control}, 2021, pp. 1--11.

\bibitem{hasanbeig2020cautious}
M.~Hasanbeig, A.~Abate, and D.~Kroening, ``Cautious reinforcement learning with logical constraints,'' \emph{arXiv preprint arXiv:2002.12156}, 2020.

\bibitem{hasanbeig2020towards}
M.~Hasanbeig, D.~Kroening, and A.~Abate, ``Towards verifiable and safe model-free reinforcement learning.''\hskip 1em plus 0.5em minus 0.4em\relax CEUR Workshop Proceedings, 2020.

\bibitem{hasanbeig2023certified}
H.~Hasanbeig, D.~Kroening, and A.~Abate, ``Certified reinforcement learning with logic guidance,'' \emph{Artificial Intelligence}, vol. 322, p. 103949, 2023.

\bibitem{elsayed2021safe}
I.~ElSayed-Aly, S.~Bharadwaj, C.~Amato, R.~Ehlers, U.~Topcu, and L.~Feng, ``Safe multi-agent reinforcement learning via shielding,'' \emph{arXiv preprint arXiv:2101.11196}, 2021.

\bibitem{carr2023safe}
S.~Carr, N.~Jansen, S.~Junges, and U.~Topcu, ``Safe reinforcement learning via shielding under partial observability,'' in \emph{Proceedings of the AAAI Conference on Artificial Intelligence}, vol.~37, no.~12, 2023, pp. 14\,748--14\,756.

\bibitem{nikou2021symbolic}
A.~Nikou, A.~Mujumdar, M.~Orlic, and A.~V. Feljan, ``Symbolic reinforcement learning for safe ran control,'' \emph{arXiv preprint arXiv:2103.06602}, 2021.

\bibitem{zhao2022safe}
Z.~Zhao, J.~Xun, X.~Wen, and J.~Chen, ``Safe reinforcement learning for single train trajectory optimization via shield sarsa,'' \emph{IEEE Transactions on Intelligent Transportation Systems}, vol.~24, no.~1, pp. 412--428, 2022.

\bibitem{alur2002alternating}
R.~Alur, T.~A. Henzinger, and et~al., ``Alternating-time temporal logic,'' \emph{Journal of the ACM}, vol.~49, no.~5, pp. 672--713, 2002.

\bibitem{liu2020modal}
Z.~Liu, L.~Xiong, Y.~Liu, Y.~Lesp{\'e}rance, R.~Xu, and H.~Shi, ``A modal logic for joint abilities under strategy commitments.'' in \emph{IJCAI}, 2020, pp. 1805--1812.

\bibitem{xiong2016strategy}
L.~Xiong and Y.~Liu, ``Strategy representation and reasoning for incomplete information concurrent games in the situation calculus.'' in \emph{IJCAI}, 2016, pp. 1322--1329.

\bibitem{liu2022lang2ltl}
J.~X. Liu, Z.~Yang, B.~Schornstein, S.~Liang, I.~Idrees, S.~Tellex, and A.~Shah, ``Lang2ltl: Translating natural language commands to temporal specification with large language models,'' in \emph{Workshop on Language and Robotics at CoRL 2022}, 2022.
\end{thebibliography}
\end{document}